\documentclass[letterpaper, 10 pt, conference]{ieeeconf}  

\IEEEoverridecommandlockouts                              

\usepackage{graphics} 
\usepackage{multirow}
\usepackage{graphicx}
\usepackage{amsmath}
\usepackage{colortbl}
\usepackage{amsfonts}
\usepackage{subfig}
\usepackage{xcolor}
\usepackage{booktabs}

\title{\LARGE \bf
RA-NeRF: Robust Neural Radiance Field Reconstruction with Accurate Camera Pose Estimation under Complex Trajectories
}

\author{Qingsong Yan$^{1}$, Qiang Wang$^{2,*}$, Kaiyong Zhao$^{3}$, Jie Chen$^{4}$, Bo Li$^{5}$, Xiaowen Chu$^{6}$ and Fei Deng$^{1}$
\thanks{$^{1}$Qingsong Yan and Fei Deng are with School of Geodesy and Geomatics, Wuhan University, China {\tt\small yanqs\_whu@whu.edu.cn, fdeng@sgg.whu.edu.cn}}%
\thanks{$^{2}$Qiang Wang is with Department of Computer Science and Technology, HIT (Shenzhen), China {\tt\small qiang.wang@hit.edu.cn}}%
\thanks{$^{3}$Kaiyong Zhao is with XGRIDS, China {\tt\small kyzhao@xgrids.com}}%
\thanks{$^{4}$Jie Chen is with Department of Computer Science, HKBU, China {\tt\small chenjie@comp.hkbu.edu.cn}}%
\thanks{$^{5}$Bo Li is with Department of Computer Science and Engineering, HKUST, China {\tt\small bli@cse.ust.hk}}%
\thanks{$^{6}$Xiaowen Chu is with Data Science and Analytics Thrust, HKUST (Guangzhou), China {\tt\small xwchu@ust.hk}}%
\thanks{$^{*}$ indicates the corresponding author.}
}

\begin{document}

\maketitle
\thispagestyle{empty}
\pagestyle{empty}


\begin{abstract}

Neural Radiance Fields (NeRF) and 3D Gaussian Splatting (3DGS) have emerged as powerful tools for 3D reconstruction and SLAM tasks. 
However, their performance depends heavily on accurate camera pose priors.
Existing approaches attempt to address this issue by introducing external constraints but fall short of achieving satisfactory accuracy, particularly when camera trajectories are complex.
In this paper, we propose a novel method, RA-NeRF, capable of predicting highly accurate camera poses even with complex camera trajectories.
Following the incremental pipeline, RA-NeRF reconstructs the scene using NeRF with photometric consistency and incorporates flow-driven pose regulation to enhance robustness during initialization and localization. Additionally, RA-NeRF employs an implicit pose filter to capture the camera movement pattern and eliminate the noise for pose estimation.
To validate our method, we conduct extensive experiments on the Tanks\&Temple dataset for standard evaluation, as well as the NeRFBuster dataset, which presents challenging camera pose trajectories.
On both datasets, RA-NeRF achieves state-of-the-art results in both camera pose estimation and visual quality, demonstrating its effectiveness and robustness in scene reconstruction under complex pose trajectories.

\end{abstract}
\section{Introduction}
\label{sec:intro}

\begin{figure}[t]

    \centering
    
    \subfloat[Nope-NeRF \cite{bian2022nope}]{\includegraphics[width=0.49\linewidth]{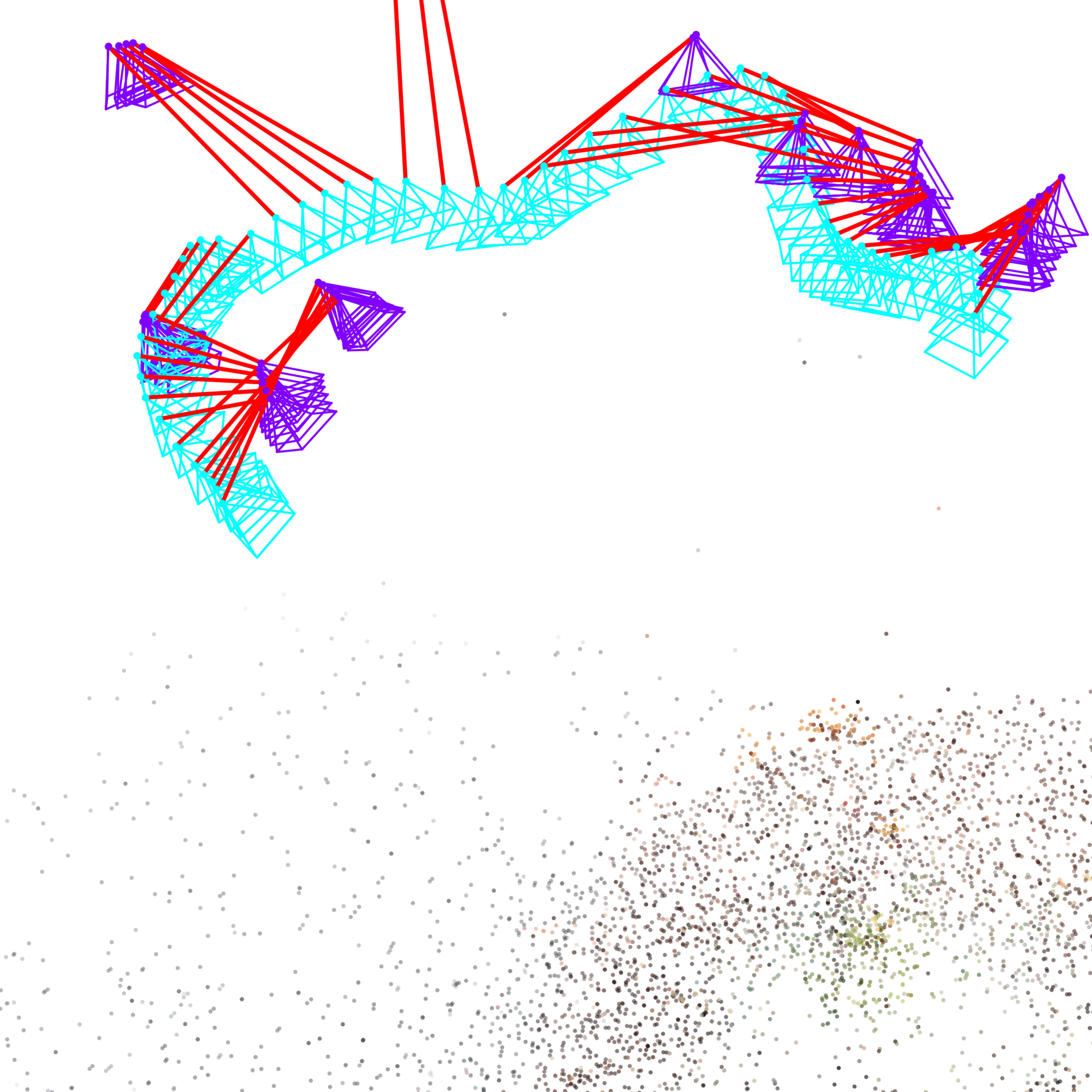}}
    \hspace{0.1em}
    \subfloat[CF-NeRF$^*$ \cite{yan2023cf}]{\includegraphics[width=0.49\linewidth]{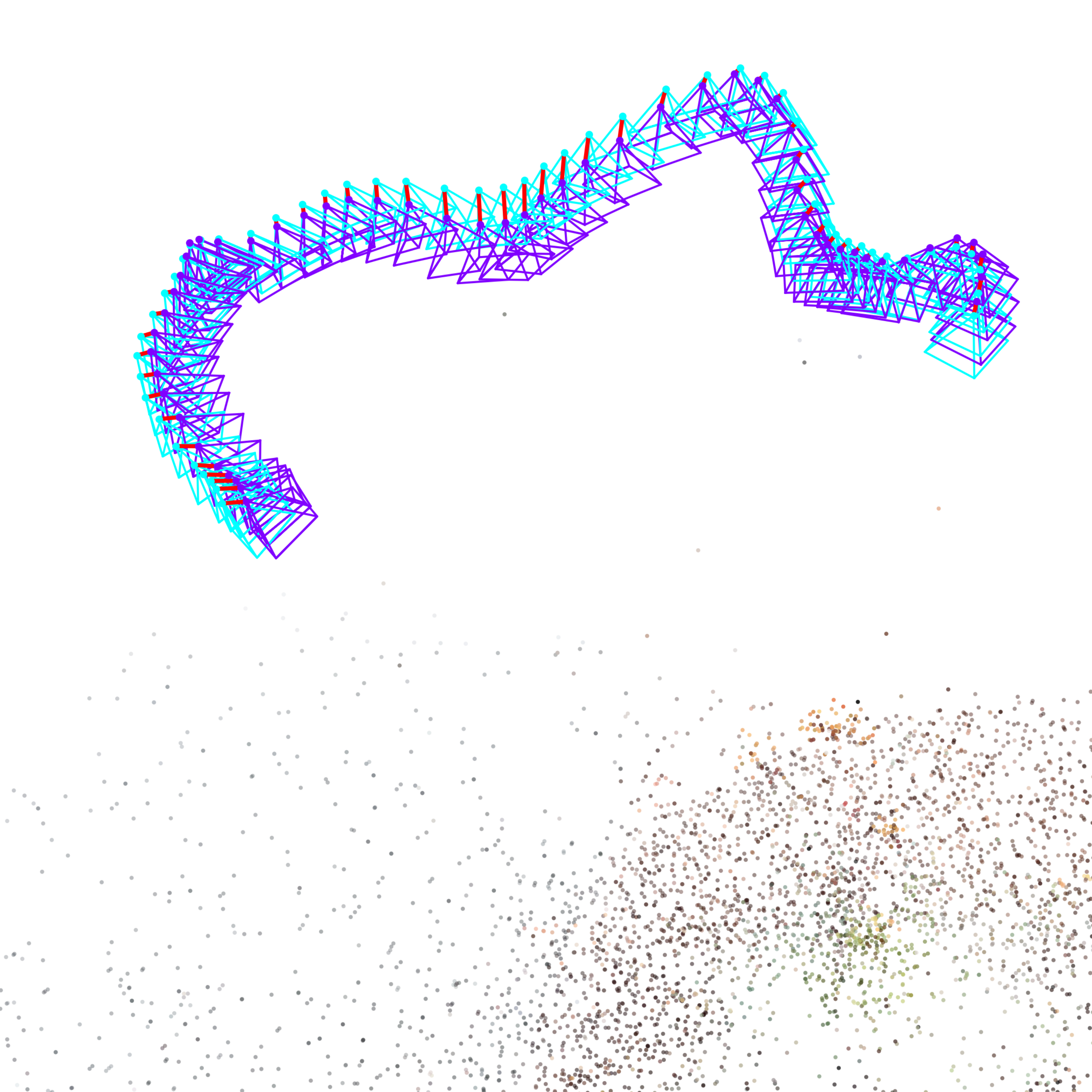}}

    \subfloat[CF-3DGS \cite{fu2024cf3dgs}]{\includegraphics[width=0.49\linewidth]{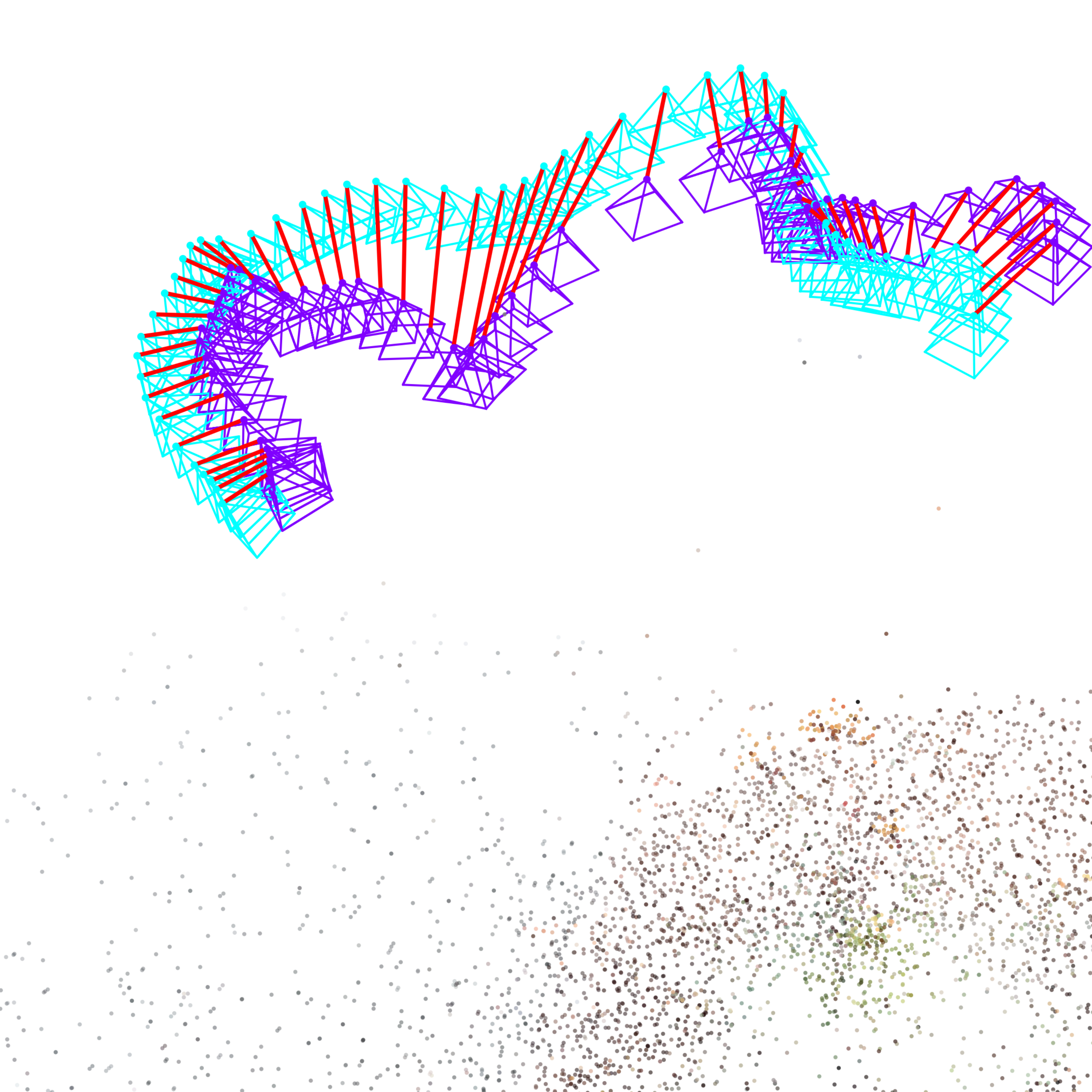}}
    \hspace{0.1em}
    \subfloat[RA-NeRF]{\includegraphics[width=0.49\linewidth]{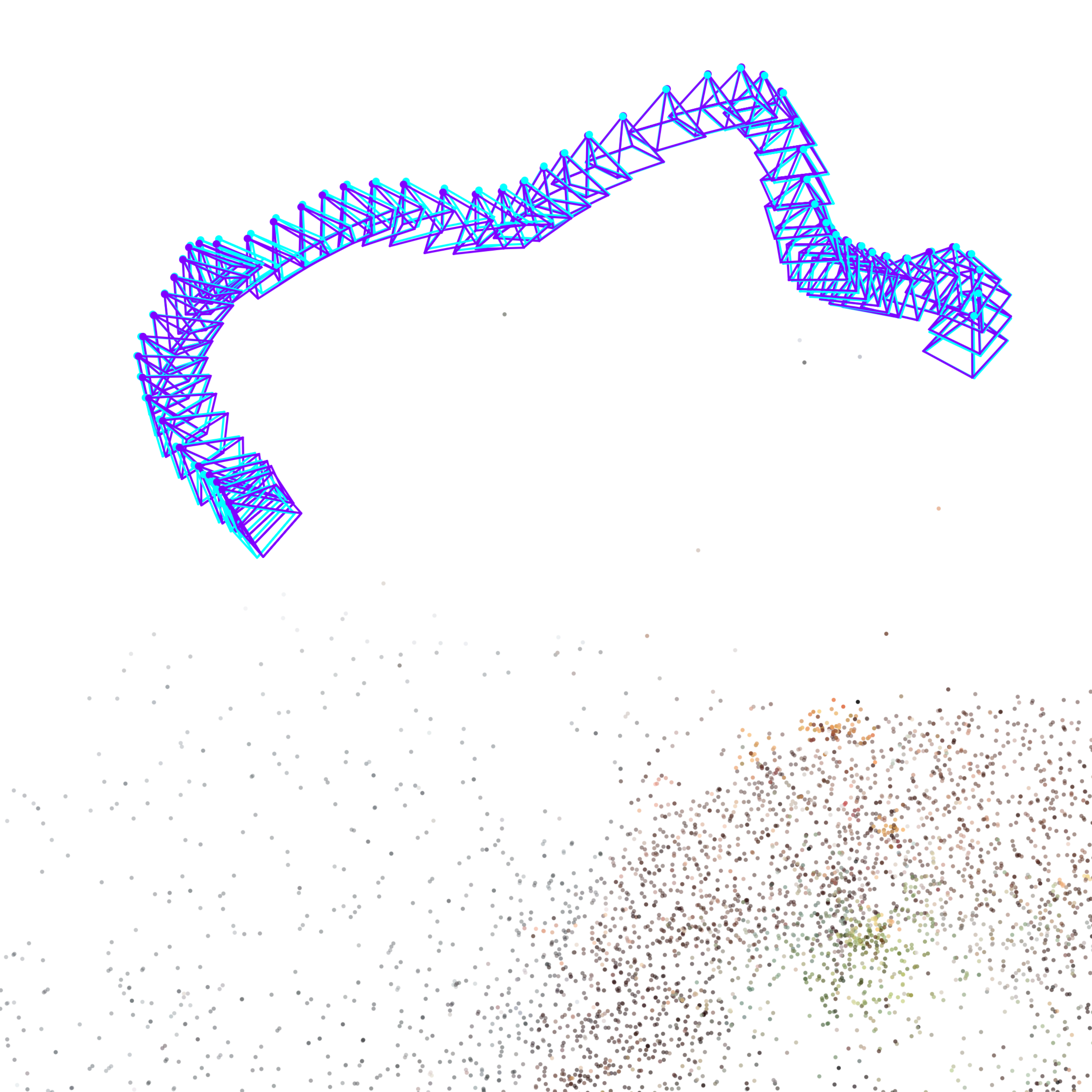}}

    \vspace{-0.5em}

    \caption{
        Visualization of estimated camera poses for one scene from the NeRFBuster dataset \cite{warburg2023nerfbusters}. 
        The \textcolor{cyan}{ground truth} camera poses are obtained using COLMAP, while the \textcolor{violet}{estimated results} are from Nope-NeRF \cite{bian2022nope}, CF-NeRF \cite{yan2023cf}, CF-3DGS \cite{fu2024cf3dgs}, and our method RA-NeRF, where CF-NeRF$^*$ means CF-NeRF does not estimate the focal length.
        The \textcolor{red}{line} indicates the error between the ground truth and the estimated poses.} 
    \label{fig:viz_poses_in_aloe}

    \vspace{-2.0em}

\end{figure}

Neural Radiance Fields (NeRF) \cite{mildenhall2021nerf} and 3D Gaussian Splatting (3DGS) \cite{kerbl3Dgaussians} have shown remarkable performance in novel view synthesis, demonstrating great potential in dynamic scene representation \cite{pumarola2021d,li2023dynibar}, surface reconstruction \cite{li2023neuralangelo,chen2024pgsr}, and large-scale reconstruction \cite{zhenxingswitch,lin2024vastgaussian}, which are fundamental problems in robotics and autonomous driving. 
However, these methods rely heavily on accurate camera poses provided by traditional tools \cite{schonberger2016structure}.

To address the above limitation, NeRFmm \cite{wang2021nerf} and BARF \cite{lin2021barf} treat camera poses as learnable variables and unify camera pose estimation and scene reconstruction.
To further improve reconstruction quality, SiRENmm \cite{ventusff2021} and GARF \cite{chng2022garf} replace the original ReLU activation function to preserve more information during the pose estimation.
Additionally, L2G-NeRF \cite{chen2023local} and LU-NeRF \cite{cheng2023lu} adopt a local-to-global strategy for camera pose and scene reconstruction.
Despite these improvements, these methods are primarily effective for forward-looking scenes, such as scenes in the LLFF dataset \cite{mildenhall2019llff}, but struggle with scenes involving complex rotations when initial poses are unavailable.

A straightforward idea is to introduce external constraints to guide pose estimation.
Nope-NeRF \cite{bian2022nope} and TD-NeRF \cite{tan2024td} use depth \cite{ranftl2021vision} but need to deal with multi-view consistency.
SC-NeRF \cite{jeong2021self}, PoRF \cite{bian2023porf}, and SPARF \cite{truong2023sparf} utilize optical flow \cite{truong2021learning} or feature matching \cite{sun2021loftr}, but they require unreliable depth from NeRF to construct the loss.
Alternatively, GNeRF \cite{meng2021gnerf} and IR-NeRF \cite{zhang2023pose} assume known image distributions.
DBARF \cite{chen2023dbarf} incorporates semantic information for pose estimation.
However, these methods are prone to local minima, especially in scenes with significant rotations.

\begin{figure*}[t]
    \centering
    
    \includegraphics[width=0.82\linewidth]{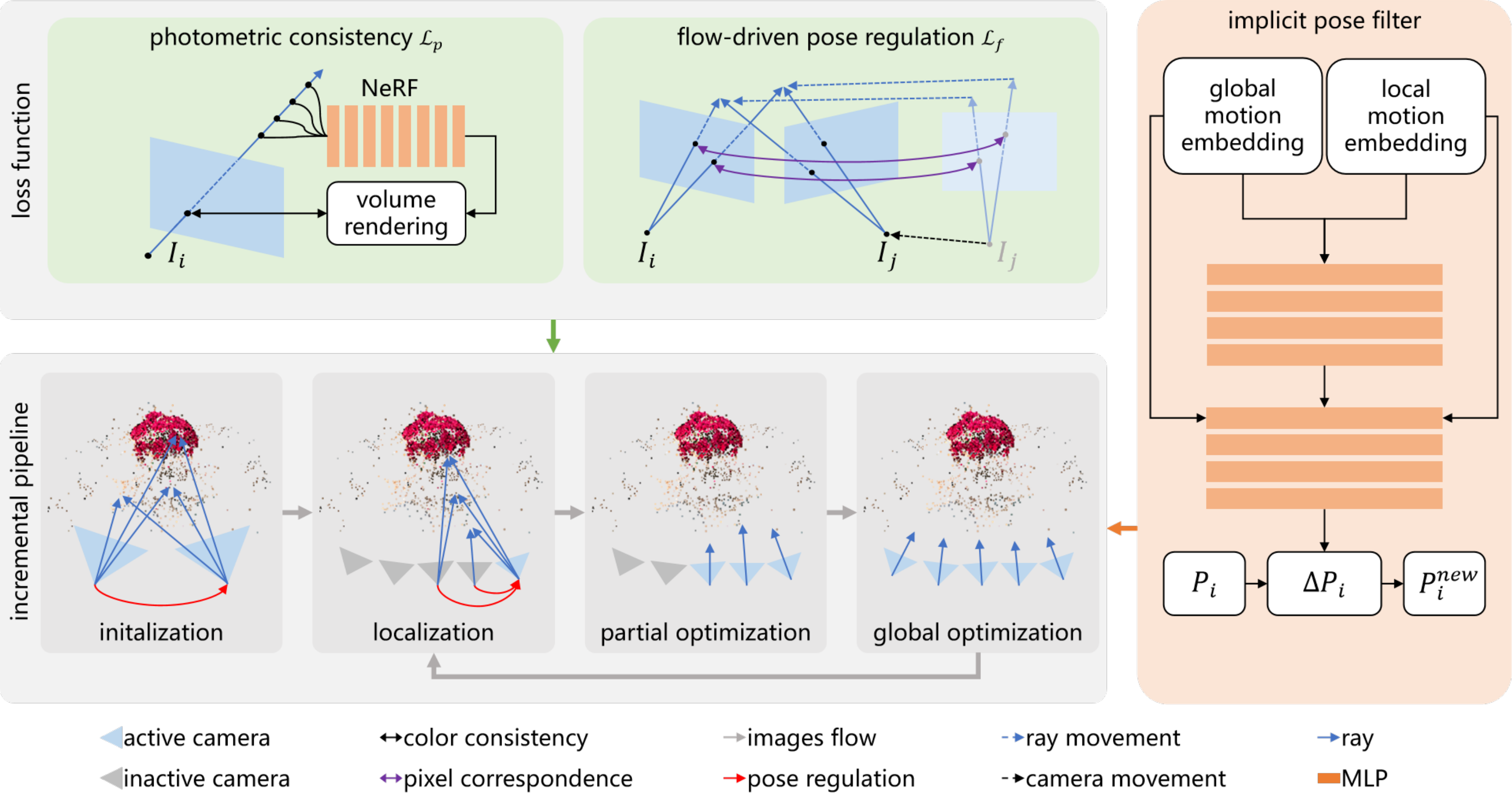}
    
    \vspace{-0.5em}

    \caption{
        The pipeline of the RA-NeRF, which uses photometric consistency to train the NeRF and camera poses, and improves reconstruction quality through flow-driven pose regulation and implicit pose filter.
        } 
    \label{fig:pipeline_of_IFNeRF}

    \vspace{-2.0em}

\end{figure*}

Inspired by incremental Structure-from-Motion (SfM) \cite{schonberger2016structure} and Simultaneous Localization and Mapping (SLAM) \cite{teed2021droid}, LocalRF \cite{meuleman2023progressively} and CF-NeRF \cite{yan2023cf} progressively reconstruct camera poses and 3D scenes from sequential images, leveraging initialization, localization, and optimization modules. 
In this process, initialization serves as the starting point, localization identifies the relative pose of newly added images, and optimization refines camera poses and the scene simultaneously. 
CT-NeRF \cite{ran2024ct} integrates optical flow \cite{edstedt2023dkm} into the incremental pipeline, whereas CF-3DGS \cite{fu2024cf3dgs} uses depth priors \cite{ranftl2021vision}. 
Nonetheless, these methods still encounter challenges in achieving high-accuracy camera poses.

To address these issues, we propose a novel method, RA-NeRF, which accurately estimates camera poses from image sequences. 
RA-NeRF follows the incremental pipeline of CF-NeRF \cite{yan2023cf} and CF-3DGS \cite{fu2024cf3dgs}.
Additionally, RA-NeRF employs flow-driven pose regulation to avoid convergence to local minima.
Unlike existing methods that directly optimize camera poses, RA-NeRF leverages the implicit pose filter and updates camera poses in SE(3).
In Fig. \ref{fig:viz_poses_in_aloe}, we compare the camera poses estimated by Nope-NeRF \cite{bian2022nope}, CF-NeRF \cite{yan2023cf}, CF-3DGS \cite{fu2024cf3dgs}, and our method, RA-NeRF. 
The results demonstrate that the camera poses estimated by RA-NeRF are the closest to the ground truth generated by COLMAP \cite{schonberger2016structure}.
Our contributions are as follows:
\begin{enumerate}

\item We propose a flow-driven pose regulation technique that significantly reduces the risk of converging to local minima during camera pose estimation.

\item We introduce an implicit pose filter capable of effectively modeling complex camera motions and mitigating noise from gradients, thereby enabling accurate pose estimation across diverse scenarios.

\item Experiments conducted on the NeRFBuster \cite{warburg2023nerfbusters, yan2023cf} and Tanks\&Temple \cite{knapitsch2017tanks, bian2022nope} datasets validate that RA-NeRF achieves state-of-the-art performance in terms of rendering quality and pose estimation on challenging trajectories with unknown camera poses.

\end{enumerate}
\section{Related Work}

\paragraph{NeRF and 3DGS} NeRF \cite{mildenhall2021nerf} utilizes a multilayer perceptron (MLP) to implicitly reconstruct the 3D scene and synthesize a high-quality novel view through volume rendering techniques. 
However, the efficiency of NeRF is limited by pixel-wise discrete sampling along each ray, motivating researchers to incorporate spatial priors for acceleration \cite{yu2021plenoctrees,muller2022instant}. 
Furthermore, NeRF struggles with unbounded scenes due to the sampling strategies \cite{barron2022mip}.
In contrast, 3DGS \cite{kerbl3Dgaussians} employs explicit 3D ellipsoids and uses rasterization for novel view synthesis. 
However, while 3DGS significantly improves reconstruction efficiency, it introduces the problem of over-reconstruction \cite{zhang2024fregs} and requires reducing model redundancy \cite{fan2023lightgaussian}.
Meanwhile, MipNeRF \cite{barron2021mip} and Mip-Splatting \cite{yu2024mip} explore how to deal with artifacts caused by resolution changes.
Moreover, \cite{jain2021putting,yang2023freenerf} attempt to maintain the reconstruction quality with sparse input images.
On the other hand, \cite{pumarola2021d,li2023dynibar} try to reconstruct the dynamic scene by using time as a separate dimension, thus presenting richer information for users. 
Considering representing a large-scale 3D scene, \cite{zhenxingswitch,lin2024vastgaussian} split the whole scene into smaller pieces and merge them during visualization.
Both NeRF and 3DGS have demonstrated impressive results in visual quality, and they also achieve significant results in surface reconstruction.
\cite{yariv2021volume,wang2021neus} reconstruct surface through modeling the relationship between transparency and the SDF, while
\cite{li2023neuralangelo} improve surface reconstruction efficiency based on Instant-NGP \cite{muller2022instant}.
In addition to this, \cite{guedon2024sugar} conduct surface reconstruction based on 3DGS.
To improve reconstruction quality, \cite{fu2022geo} explore how to reconstruct the surface under sparse views, and \cite{chen2024pgsr} introduces external constraints.
Despite their successes, these methods depend heavily on accurate camera poses obtained from traditional methods \cite{schonberger2016structure}, which splits the reconstruction pipeline into two distinct stages.

\paragraph{Camera Pose Estimation} Building on advancements in SfM \cite{ schonberger2016structure} and SLAM \cite{teed2021droid}, several works have attempted to estimate camera poses using NeRF and 3DGS.
The simplest application of this idea is to determine the pose of a new image given a pre-trained NeRF \cite{yen2021inerf}.
Moreover, NeRFmm \cite{wang2021nerf} directly takes camera poses as learnable parameters, and is able to reconstruct both the scene and camera poses in forward-looking scenes.
Based on this, SiRENmm \cite{ventusff2021} and GARF \cite{chng2022garf} replace ReLU in NeRF with functions such as sin or gaussian, to improve the quality of the pose estimation, while BARF \cite{lin2021barf} uses position embedding with the coarse-to-fine strategy. L2G-NeRF \cite{chen2023local} and LU-NeRF \cite{cheng2023lu} use the local-to-global strategy.
However, the above methods only rely on the photometric consistency, and the pose quality is still not high.
\cite{boss2022samurai,zhang2023pose} assume that the distribution of image poses is known, while \cite{jeong2021self,bian2022nope,truong2023sparf} use the prior information such as the depth or optical flow.
In recent years, LocalRF \cite{meuleman2023progressively} and CF-NeRF \cite{yan2023cf} use incremental pipelines to reconstruct not only forward-looking but also rotating scenes.
CT-NeRF \cite{ran2024ct} uses optical flow, and CF-3DGS \cite{fu2024cf3dgs} uses the depth as constraints during the incremental pipeline.
In addition, \cite{tian2023mononerf} focus on generalization representation but do not focus on quality of camera poses. 
In conclusion, although the above methods achieve good results in reconstruction, they either only handle forward-looking scenes or require initial poses or prior information, and make it challenging to obtain high-quality camera poses in the case of complex motion trajectories.
To address this problem, this paper proposes RA-NeRF, which can reconstruct high-accurate camera poses under the incremental pipeline using flow-driven pose regularization of camera poses and noise suppression by implicit pose filter.

\section{Method}
In this section, we begin by introducing the fundamentals of photometric consistent. 
Next, we present the core techniques of RA-NeRF, flow-driven pose regulation and implicit pose filter, that enhance the accuracy of pose estimation. 
Finally, we describe the incremental learning pipeline of RA-NeRF, which integrates these components to simultaneously reconstruct neural implicit representations and poses from unposed images. 
The overall framework of RA-NeRF is illustrated in Fig. \ref{fig:pipeline_of_IFNeRF}.

\subsection{Photometric Consistency with NeRF}

Given a collection of images $\mathcal{I} = {I_1, I_2, I_3, ..., I_N}$, RA-NeRF needs to reconstruct the NeRF model $\mathcal{F}$ and estimate camera poses $\mathcal{P} = {P_1, P_2, P_3, ..., P_N}$, where $P_i$ includes camera translation $T_i$ and camera rotation $R_i$ for the image $I_i$.
Following BARF \cite{lin2021barf}, CF-NeRF \cite{yan2023cf}, and SPARF \cite{truong2023sparf}, $\mathcal{F}$ does not use the positional encoding and the fine network with hierarchical volume sampling during the pose estimation. 
$\mathcal{F}$ consists of 8 density MLP layers of width 256, using the ReLU activation function with inputs of 3D points and view directions.
For a pixel $p_i$ in $I_i$ with the color $c_{p_i}$, the ray $r_{p_i}$ can be obtained through $p_i \sim K( R_i P + T_i )$, according to the inverse process of projection.


To render the color $\hat{c}_{p_i}$ of $p_i$, $\mathcal{F}$ first uniformly samples points along $r_{p_i}$ with a given nearest depth $d_{min}$ and farthest depth $d_{max}$, then predicts color $c$ and density $\sigma$ of these points, and finally generates $\hat{c}_{p_i}$ through volume rendering.
As the volume rendering is differentiable and treating $\mathcal{F}$ and $\mathcal{P}$ as learnable parameters, RA-NeRF can reconstruct $\mathcal{F}$ and estimate $\mathcal{P}$ simultaneously by minimizing the photometric inconsistency as Eq. \ref{eq:photometric_loss} shows.

\begin{equation}
    \label{eq:photometric_loss} 
    \mathcal{L}_p = \sum_{p \in \mathcal{I}}{||c_p-\hat{c}_p||}
\end{equation}

\subsection{Flow-driven Pose Regulation}
\label{sec:flow_constrain}

However, the gradient of photometric consistency is not smooth, and can easily fall into local minima.
Nope-NeRF \cite{bian2022nope} and CF-3DGS \cite{fu2024cf3dgs} use depth to offer geometric cues for pose estimation but are facing issues with multi-view scale inconsistency.
Meanwhile, SC-NeRF \cite{jeong2021self}, SPARF \cite{truong2023sparf}, PoRF \cite{bian2023porf}, and CT-NeRF \cite{ran2024ct}, on the other hand, use optical flow to constrain the direction of convergence of camera poses, but either require an initial pose to construct the epipolar loss or depth to construct the re-projection error, and also face the problem of ambiguity as pointed out by LU-NeRF \cite{cheng2023lu}.
To address these issues, RA-NeRF introduces a flow-driven pose regulation based on optical flow, which calculates relative camera poses from the optical flow of an image pair using multi-view geometry and directly regulates corresponding camera poses. 

Specifically, given an image pair $< I_i, I_j > \in \mathcal{I}$, RA-NeRF first computes the optical flow $M_{ij} = \{m_i, m_j\}$ by PDC-Net \cite{truong2021learning}, where $p_i$ and $p_j$ are pixels on $I_i$ and $I_j$, respectively. 
Then, RA-NeRF calculates the fundamental matrix $F_{ij}$ for $< I_i, I_j >$ using the eight-point method \cite{hartley1997defense} and converts $F_{ij}$ to the essential matrix $E_{ij}$ via $E_{ij} = K_j^{-1} F_{ij} K_i^{-1}$. 
Finally, RA-NeRF obtains the relative camera pose $\hat{R}_{ij}$ and $\hat{T}_{ij}$ by decomposing $E_{ij}$ using SVD, and selects the valid solution by ensuring that the triangulated depth of matched pixels is positive.
Meanwhile, RA-NeRF calculates the relative camera pose $R_{ij}=R_jR_i^T, T_{ij}=T_j-R_{ij}T_i$ between $I_i$ and $I_j$ from $P_i$ and $P_j$.

\begin{figure}[t]

    \centering
    
    \subfloat[]{\includegraphics[width=0.4\linewidth]{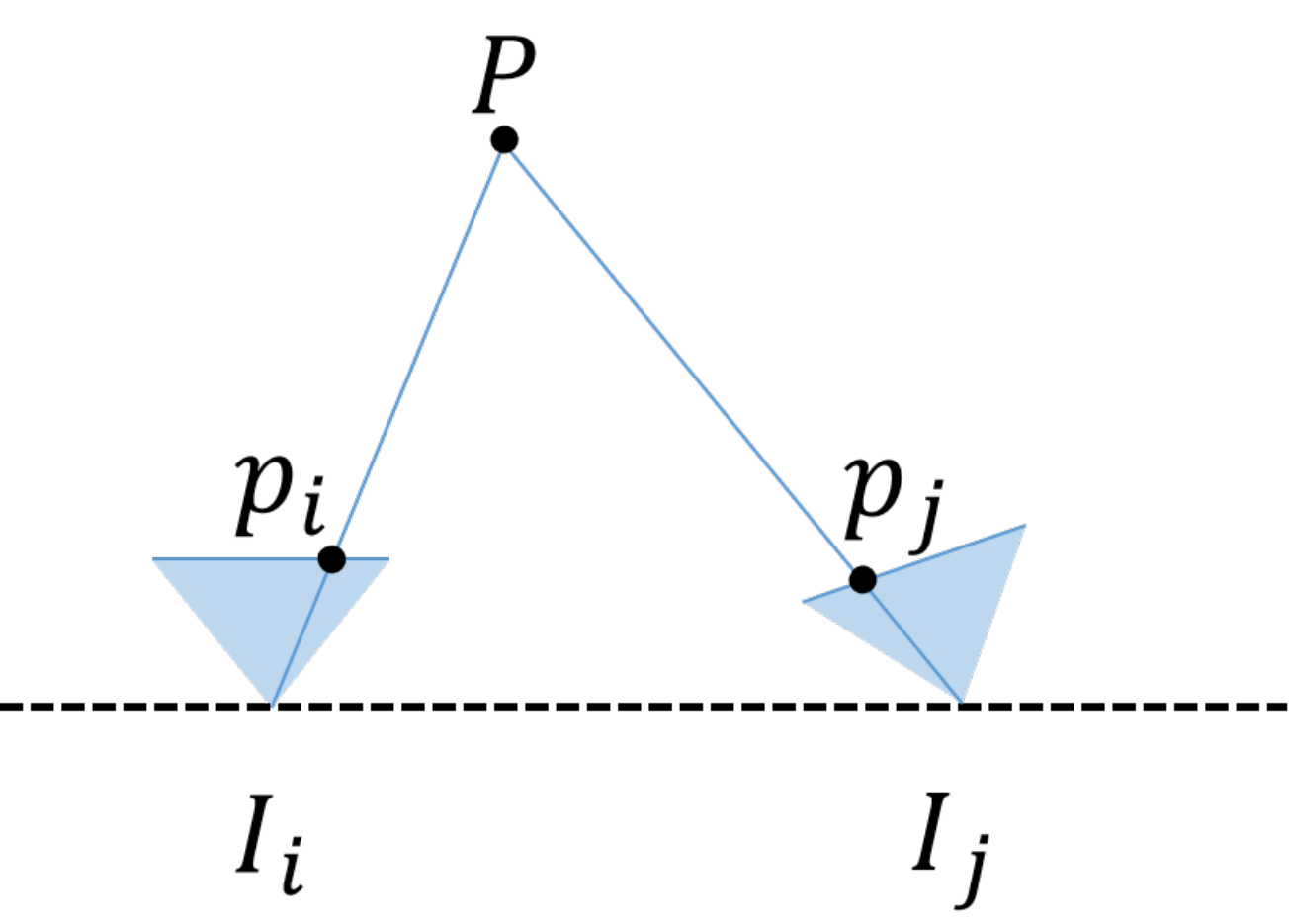}}
    \hspace{0.1em}
    \subfloat[]{\includegraphics[width=0.4\linewidth]{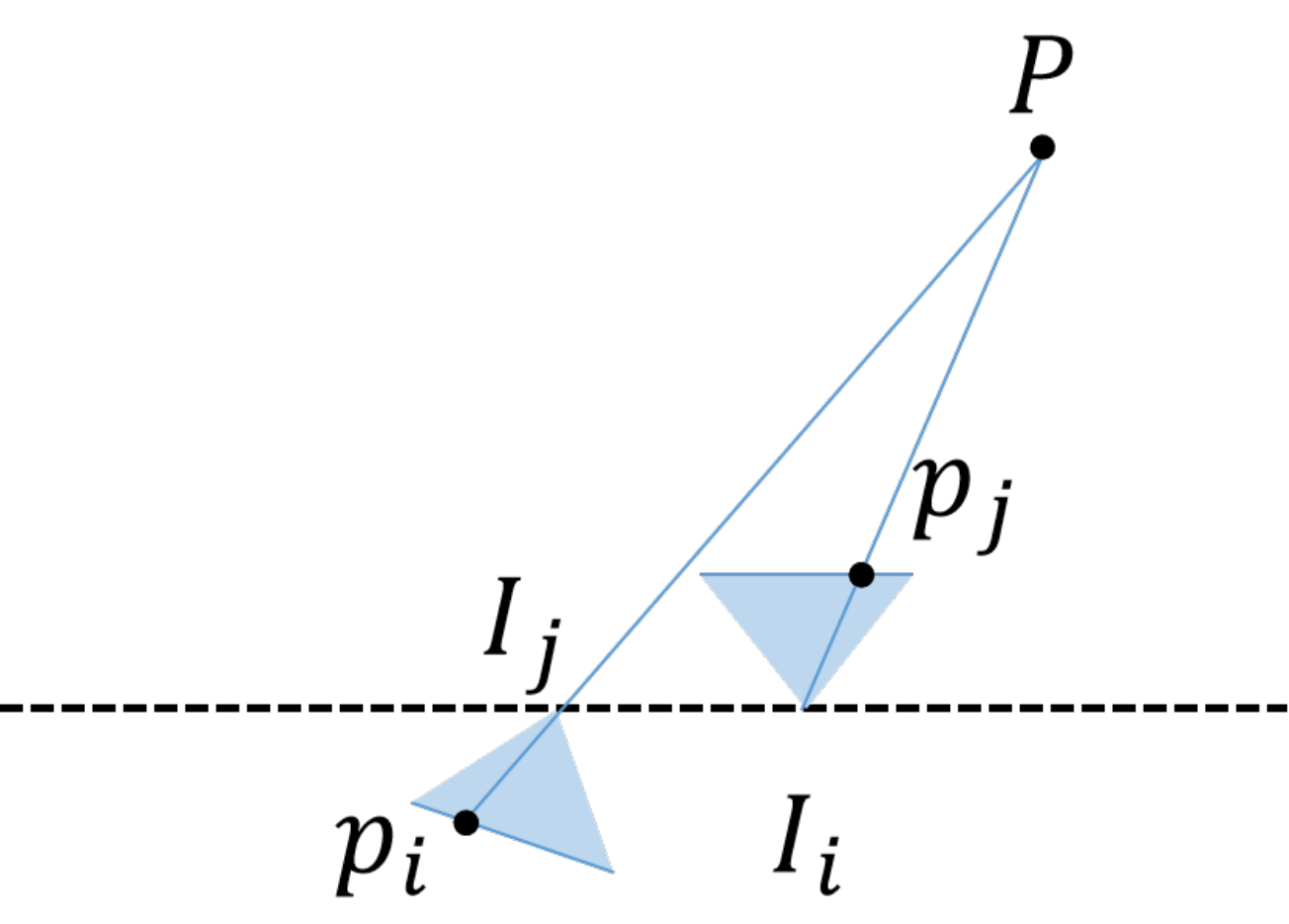}}

    \subfloat[]{\includegraphics[width=0.4\linewidth]{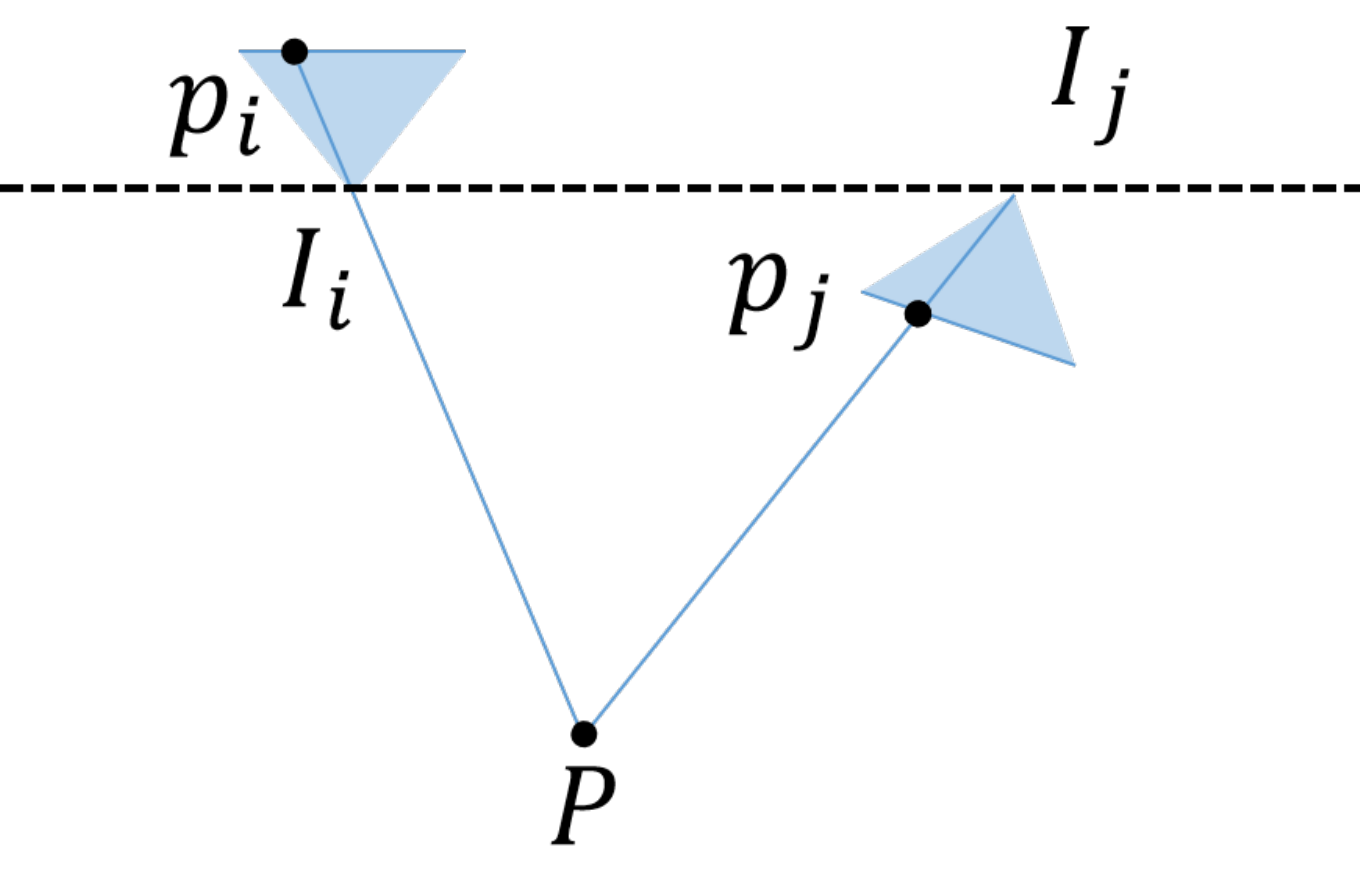}}
    \hspace{0.1em}
    \subfloat[]{\includegraphics[width=0.4\linewidth]{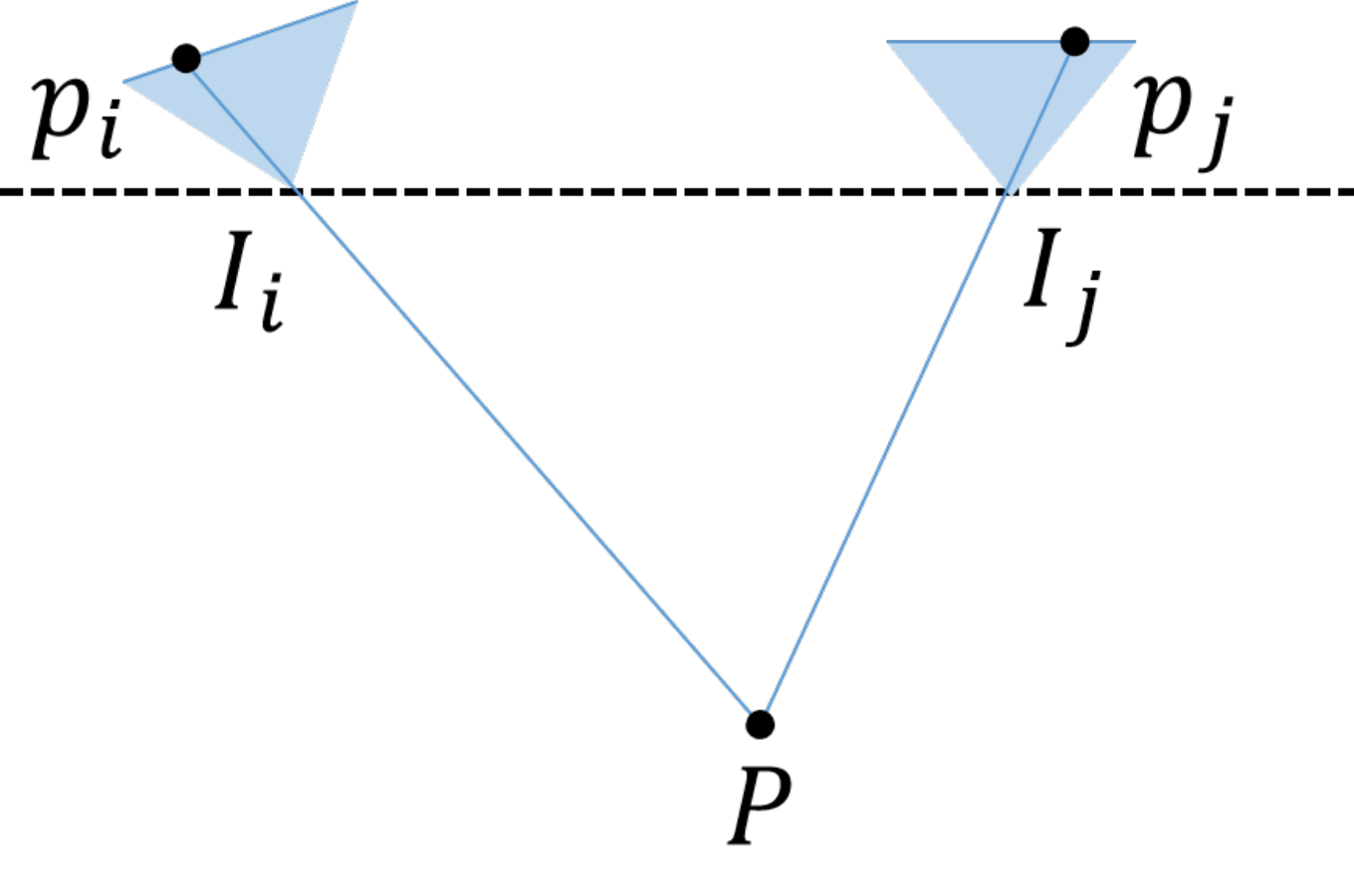}}

    \vspace{-0.5em}

    \caption{
        Given the essential matrix $E_{ij}$ of matched pixels $p_i, p_j$, we can decompose $E_{ij}$ with four possible poses through multi-view geometry \cite{hartley2003multiple}, which can mislead projected ray distance  \cite{jeong2021self} and Sampson distance \cite{fathy2011fundamental}.
        } 
    \label{fig:essential_to_rt}

    \vspace{-1.5em}

\end{figure}

It is important to note that $E_{ij}$ can yield four potential relative camera poses, as shown in Fig. \ref{fig:essential_to_rt}, where (a) and (d) are ambiguities pointed out by LU-NeRF \cite{cheng2023lu}.
Projected ray distance \cite{jeong2021self} and Sampson distance \cite{fathy2011fundamental} cannot avoid this ambiguity and cannot calculate $E_{ij}$ when $P_i=P_j$ as $T_{ij} = 0$.
However, directly calculating the relative camera pose does not suffer from these problems. 

When the estimated camera poses $\mathcal{P}$ are close to the ground truth, the differences between $(R_{ij}, T_{ij})$ and $(\hat{R}_{ij}, \hat{T}_{ij})$ should be minimal.
However, as RA-NeRF estimates camera poses from scratch, the initial difference between $(R_{ij}, T_{ij})$ and $(\hat{R}_{ij}, \hat{T}_{ij})$ is substantial.
To mitigate this, RA-NeRF regulates the estimated camera poses by directly minimizing the difference between these relative camera poses, as shown in Eq. \ref{eq:relative_loss}.
Notably, due to the absence of an absolute scale for the relative translation, RA-NeRF only regulates the direction of the translation vector.

\begin{equation}
    \label{eq:relative_loss}
    \mathcal{L}_{f} = | \frac{T_{ij}}{||T_{ij}||} - \frac{\hat{T}_{ij}}{||\hat{T}_{ij}||} |  + \lambda_{r}|R_{ij}-\hat{R}_{ij}| 
\end{equation}

Due to the large number of matching pixels between $I_i$ and $I_j$, to reduce the computational complexity, RA-NeRF randomly selects a subset of the matching pixels in each iteration to compute the relative camera pose, using a strategy similar to RANSAC to improve efficiency and robustness.

\subsection{Implicit Pose Filter}
\label{sec:implicit_pose}

Based on photometric consistency and flow-driven pose regulation, RA-NeRF can estimate camera poses from scratch.
Similar to the importance of selecting a suitable way to estimate the rotation matrix during point cloud alignment \cite{chen2022projective}, RA-NeRF updates camera poses in the Lie algebra $\mathfrak{se}(3)$, which represents the motion by a 6-dimensional vector \cite{lin2021barf,chen2023local}, and set it to learnable parameters.
To estimate the camera pose of $I_i \in \mathcal{I}$, RA-NeRF starts from the initial pose $P_i$ of $I_i$, and estimates the update values $\Delta_i = (t_i,w_i)$, which can be converted to $\Delta P_{i}$ by Eq. \ref{eq:se3_to_SE3}, where $e$ is the exponential map, $w_i^{[\times]}$ is the skew symmetric matrix generated by $w_i$, $\theta_i=|w_i|$, and $V=I+w_i^{[\times]}(1-cos\theta_i)/\theta_i^2+(w_i^{[\times]})^2(\theta_i-sin\theta_i)/\theta_i^3$.
Finally, RA-NeRF obtains the new pose $P_i^{new} = \Delta P_{i} \cdot P_{i}$.

\begin{equation}
    \label{eq:se3_to_SE3} 
    \Delta P_{i} = \begin{bmatrix}
        e^{w_i^{[\times]}} & V_it_i \\
        0 & 1
    \end{bmatrix}
\end{equation}

If photometric consistency and flow-driven pose regulation were not affected by noise and provided consistently correct gradients for estimating $\Delta_i$, then $P_i^{new}$ would be iteratively close to the ground truth pose.
However, there are unreliabilities in both the imaging process and the optical flow, and ignoring these noises will lead to inaccurate results.

To address this problem, RA-NeRF uses the implicit pose filter to deal with noise in gradients, which consists of an 8-layer MLP with a residual connection in the middle layer to enhance the learning capability, as Fig. \ref{fig:pipeline_of_IFNeRF} shows.
The input of the implicit pose filter consists of a learnable global motion embedding of shape $(1,256)$, and a learnable local motion embedding of shape $(N,256)$, where $N$ is the number of images.
The output of the implicit pose filter is $\Delta$ of shape $(N,6)$.
The global motion embedding represents the overall motion pattern across all images, while the local motion embedding captures the unique motion state of each image. 

\begin{figure*}[tp]

    \centering

    \hspace{0.18\linewidth}
    \hspace{0.1em}
    {\includegraphics[width=0.18\linewidth]{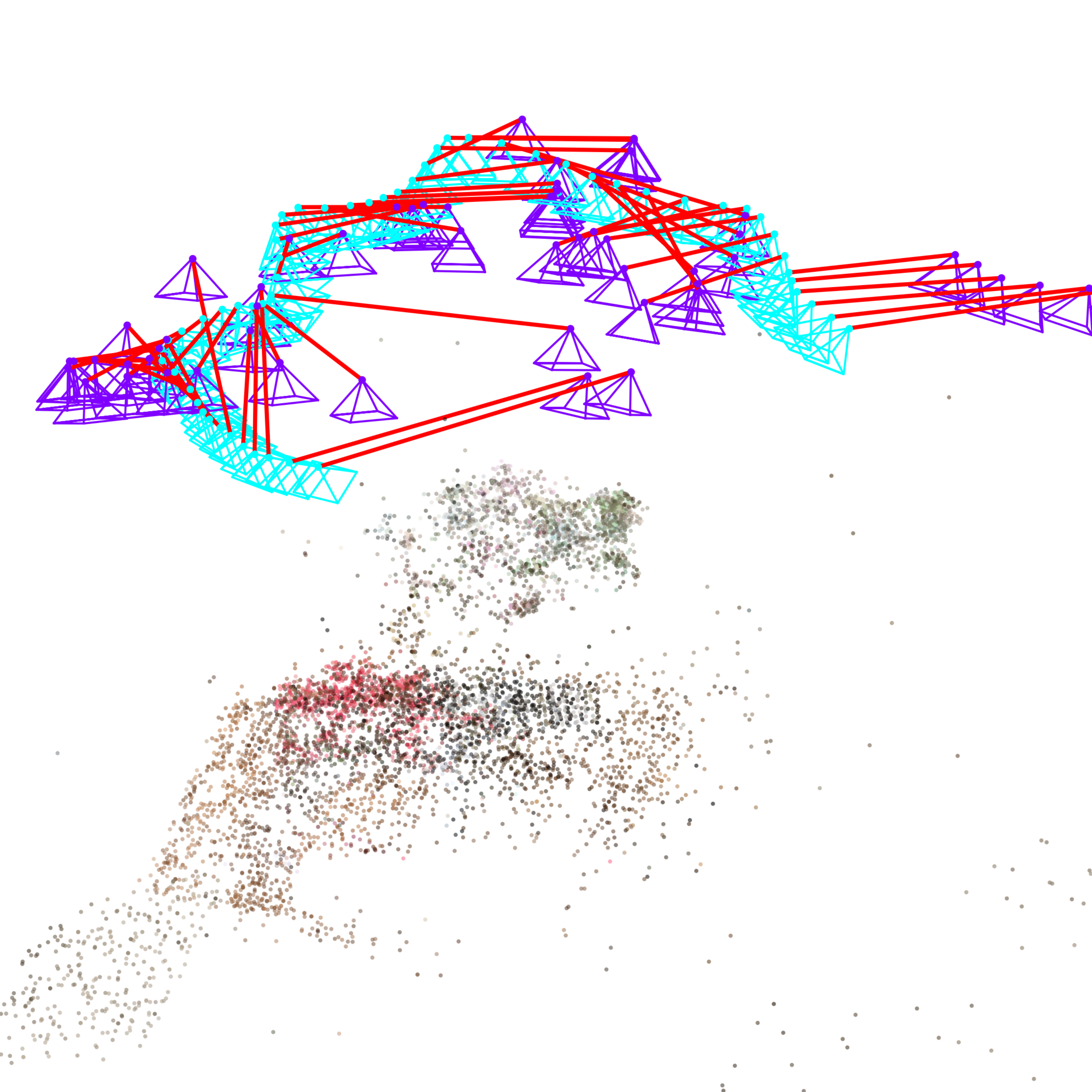}}
    \hspace{0.1em}
    {\includegraphics[width=0.18\linewidth]{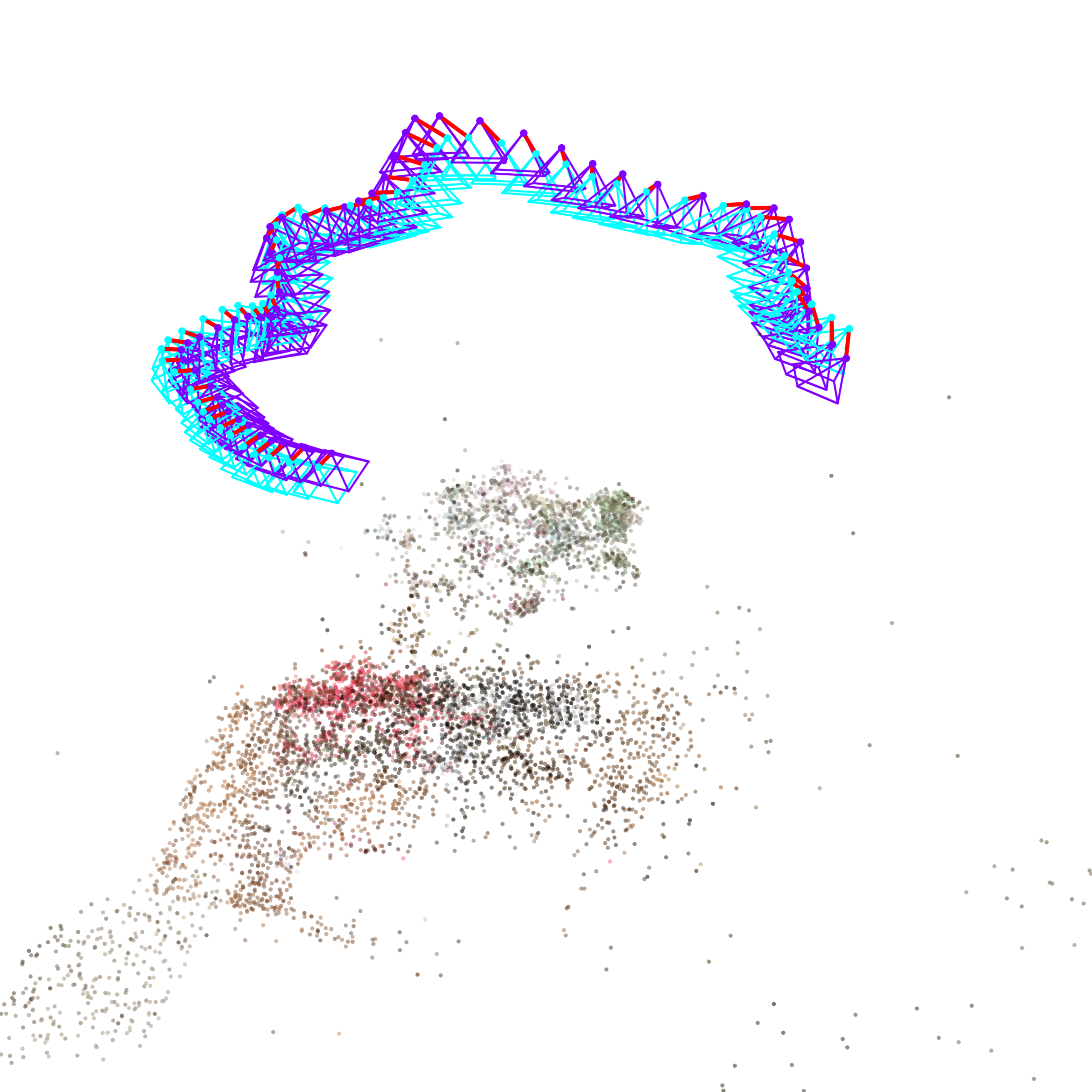}}
    \hspace{0.1em}
    {\includegraphics[width=0.18\linewidth]{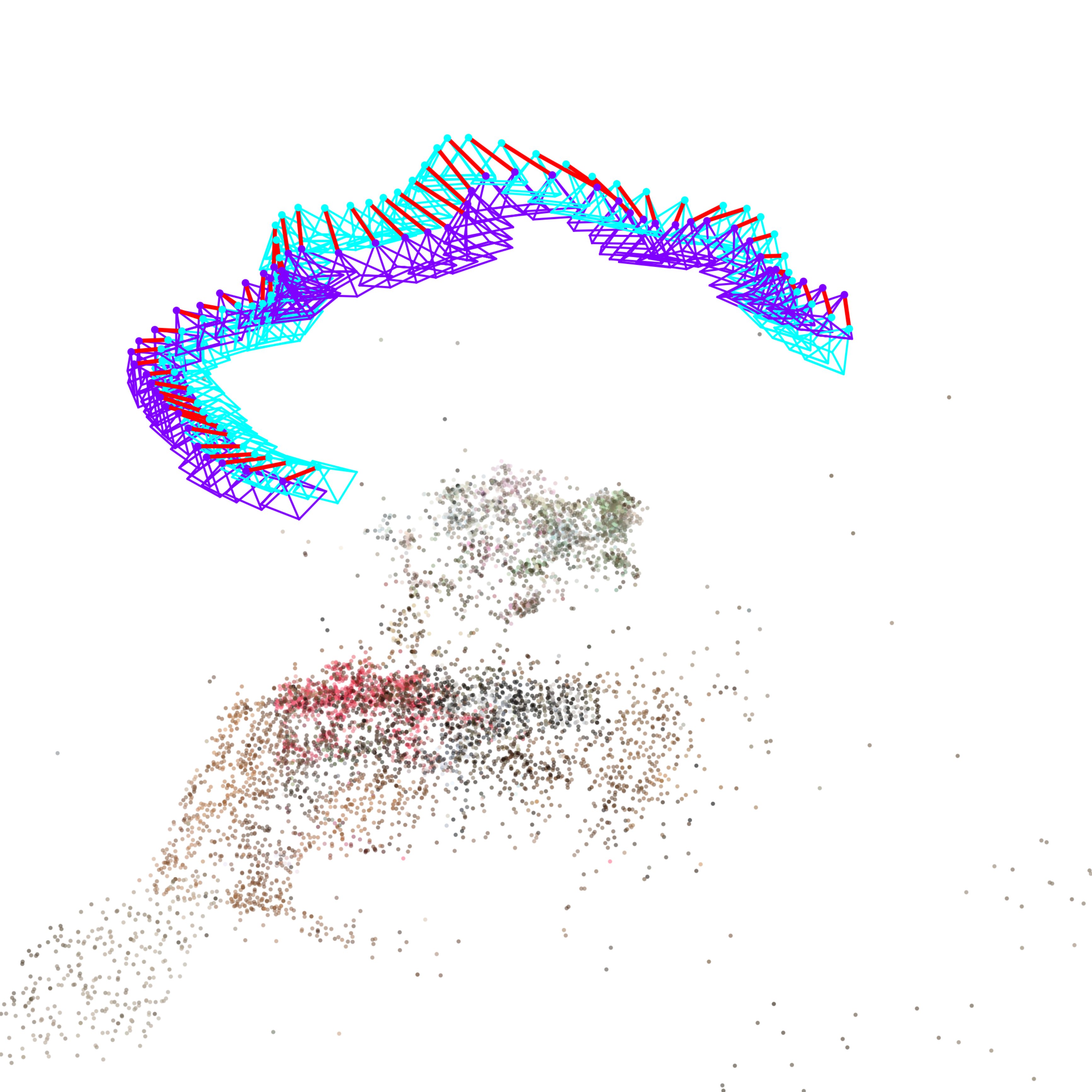}}
    \hspace{0.1em}
    {\includegraphics[width=0.18\linewidth]{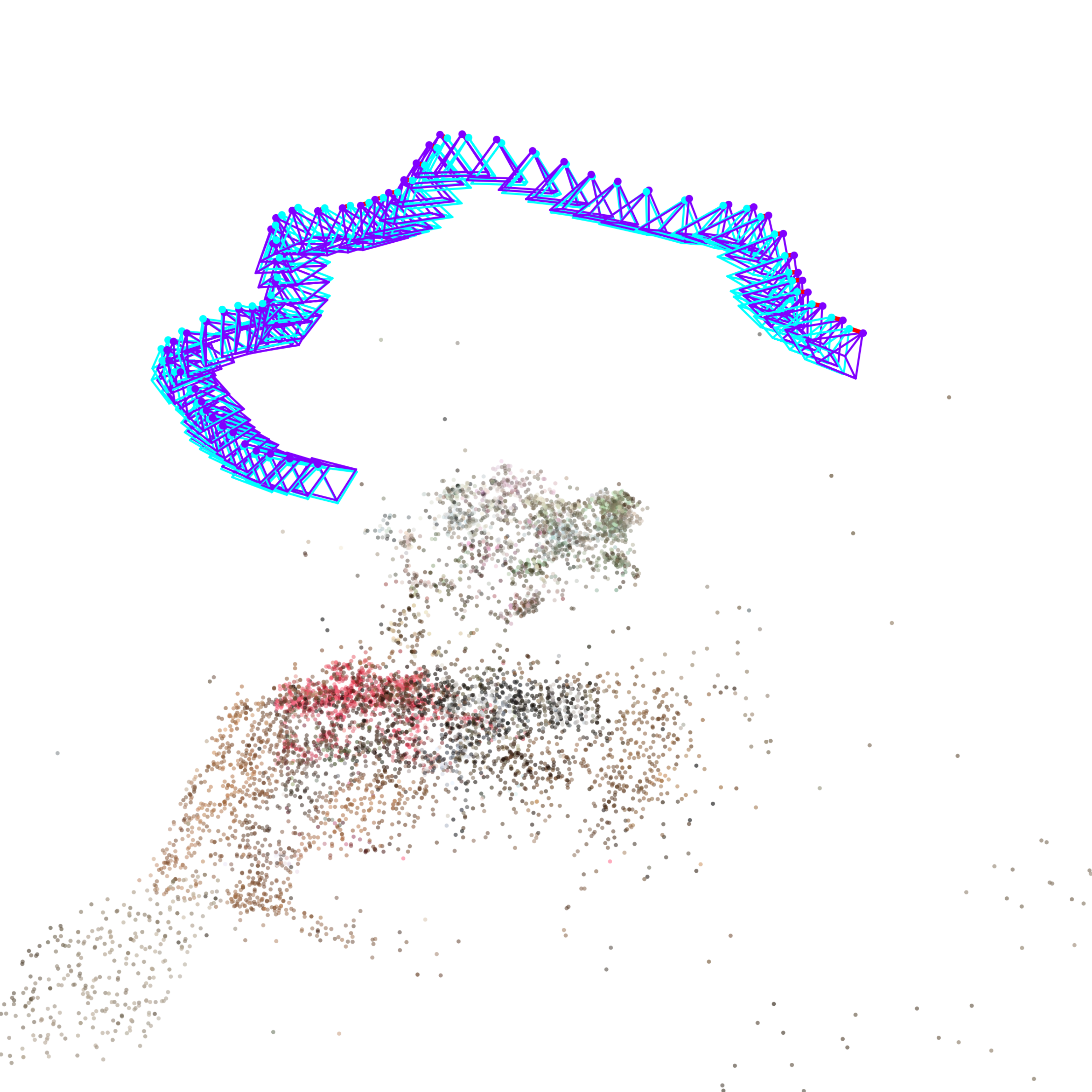}}

    \subfloat[RGB]{\includegraphics[height=0.18\linewidth, angle=-90]{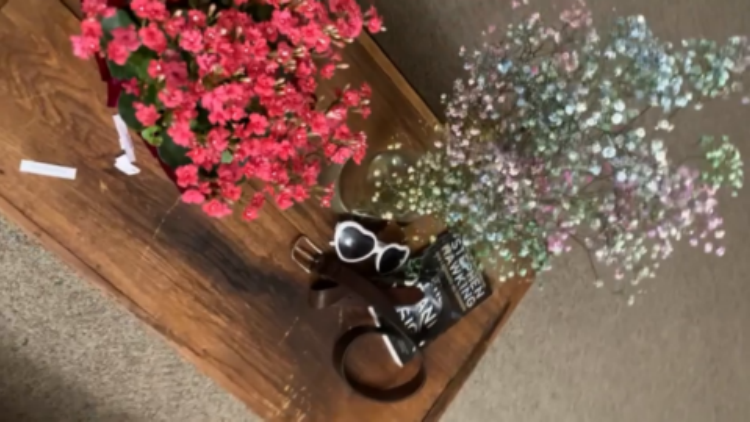}}
    \hspace{0.1em}
    \subfloat[Nope-NeRF \cite{bian2022nope}]{\includegraphics[height=0.18\linewidth, angle=-90]{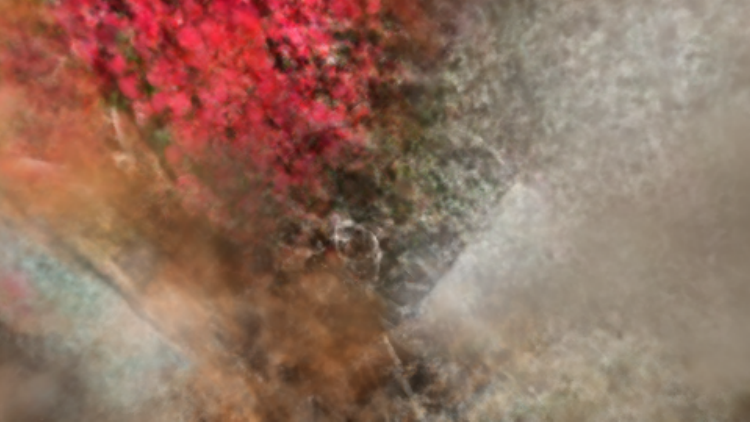}}
    \hspace{0.1em}
    \subfloat[CF-NeRF$^*$ \cite{yan2023cf}]{\includegraphics[height=0.18\linewidth, angle=-90]{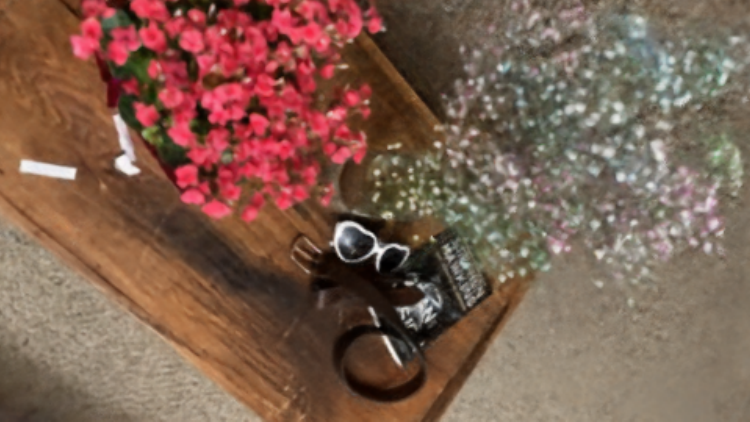}}
    \hspace{0.1em}
    \subfloat[CF-3DGS \cite{fu2024cf3dgs}]{\includegraphics[height=0.18\linewidth, angle=-90]{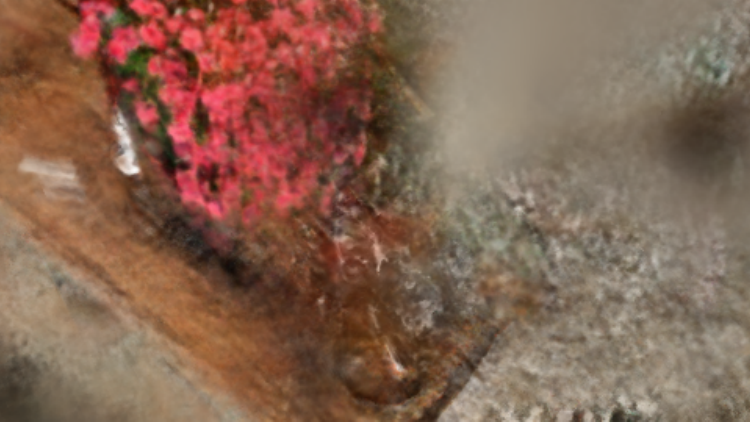}}
    \hspace{0.1em}
    \subfloat[RA-NeRF]{\includegraphics[height=0.18\linewidth, angle=-90]{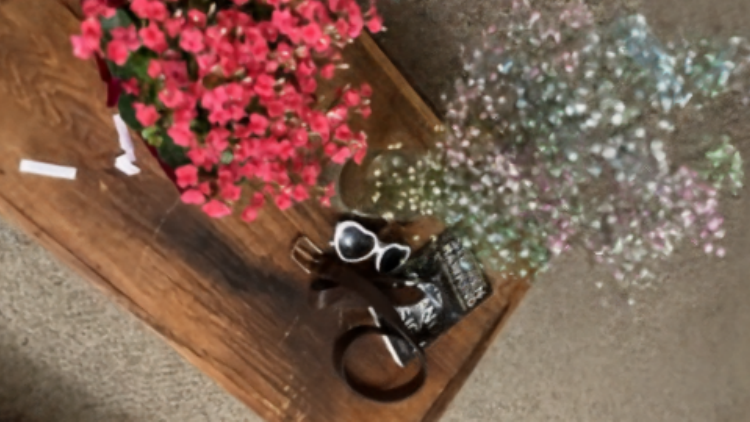}}

    \hspace{0.18\linewidth}
    \hspace{0.1em}
    {\includegraphics[width=0.18\linewidth]{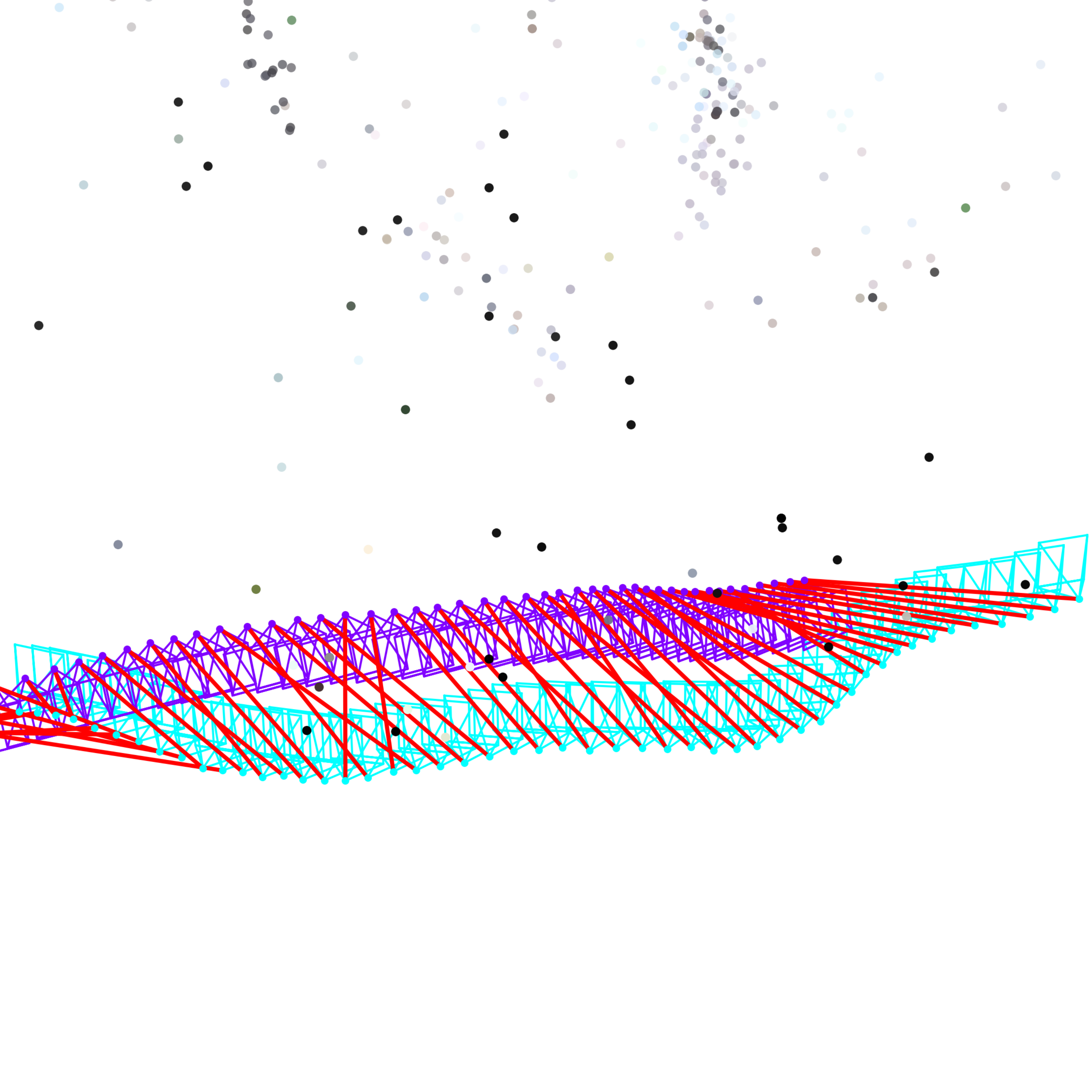}}
    \hspace{0.1em}
    {\includegraphics[width=0.18\linewidth]{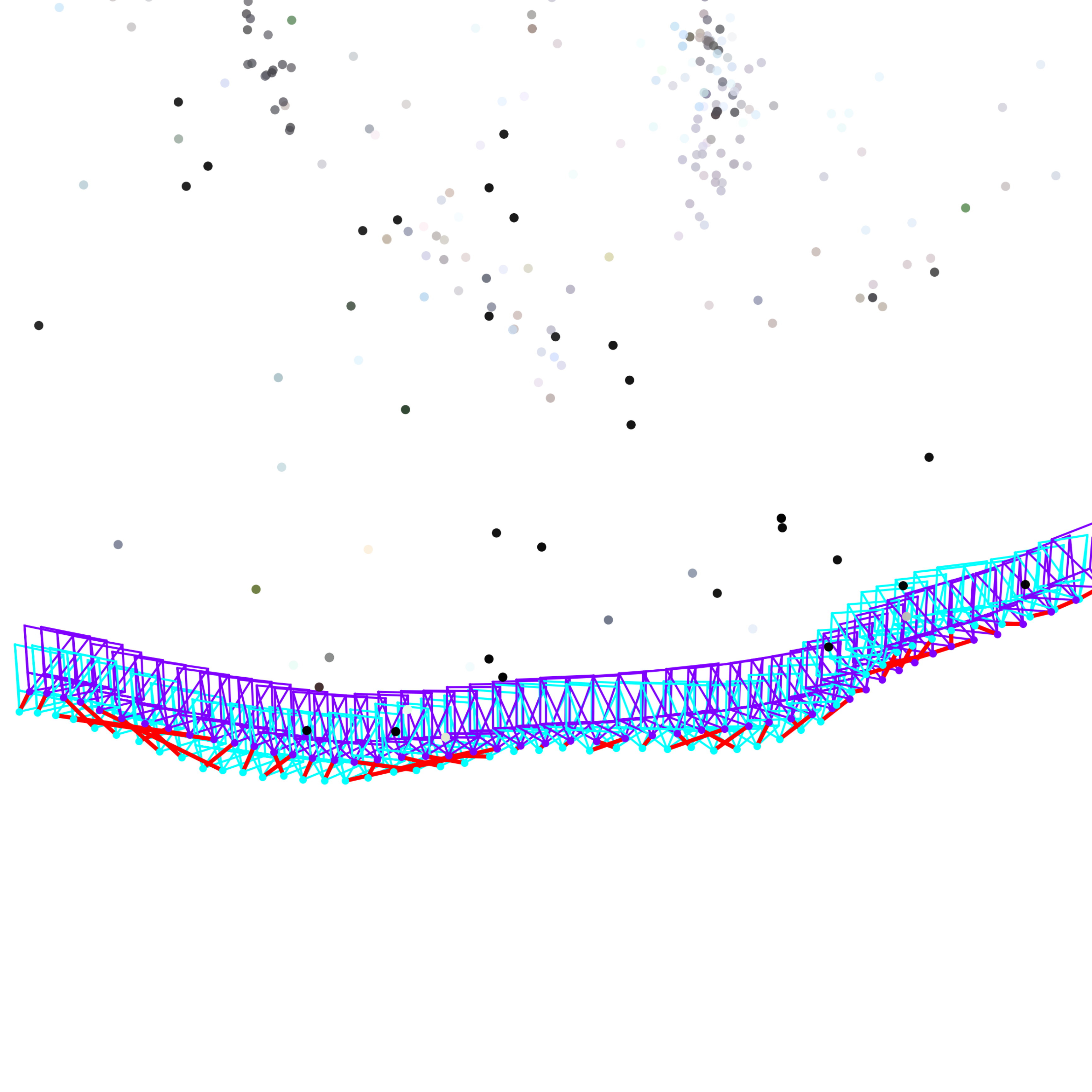}}
    \hspace{0.1em}
    {\includegraphics[width=0.18\linewidth]{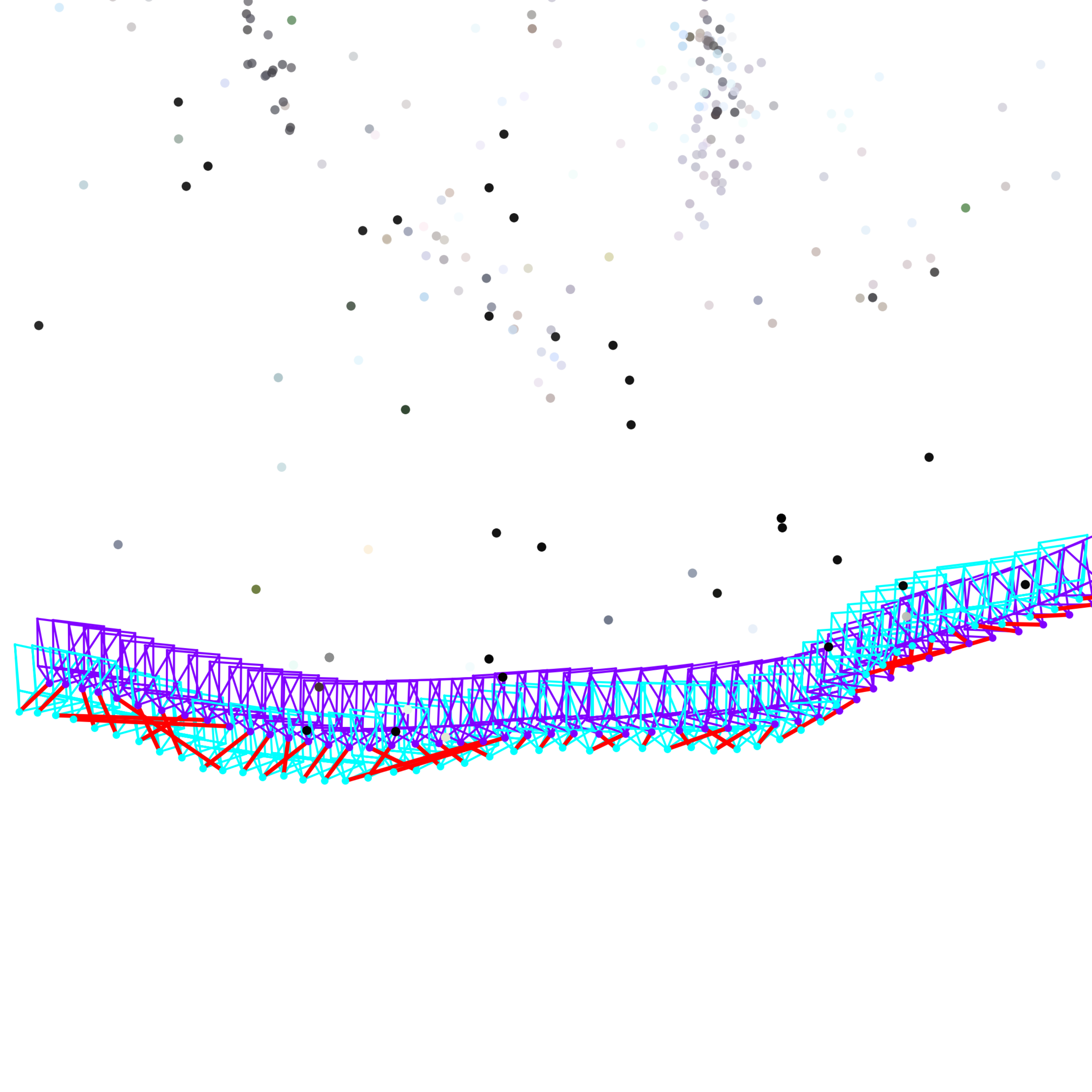}}
    \hspace{0.1em}
    {\includegraphics[width=0.18\linewidth]{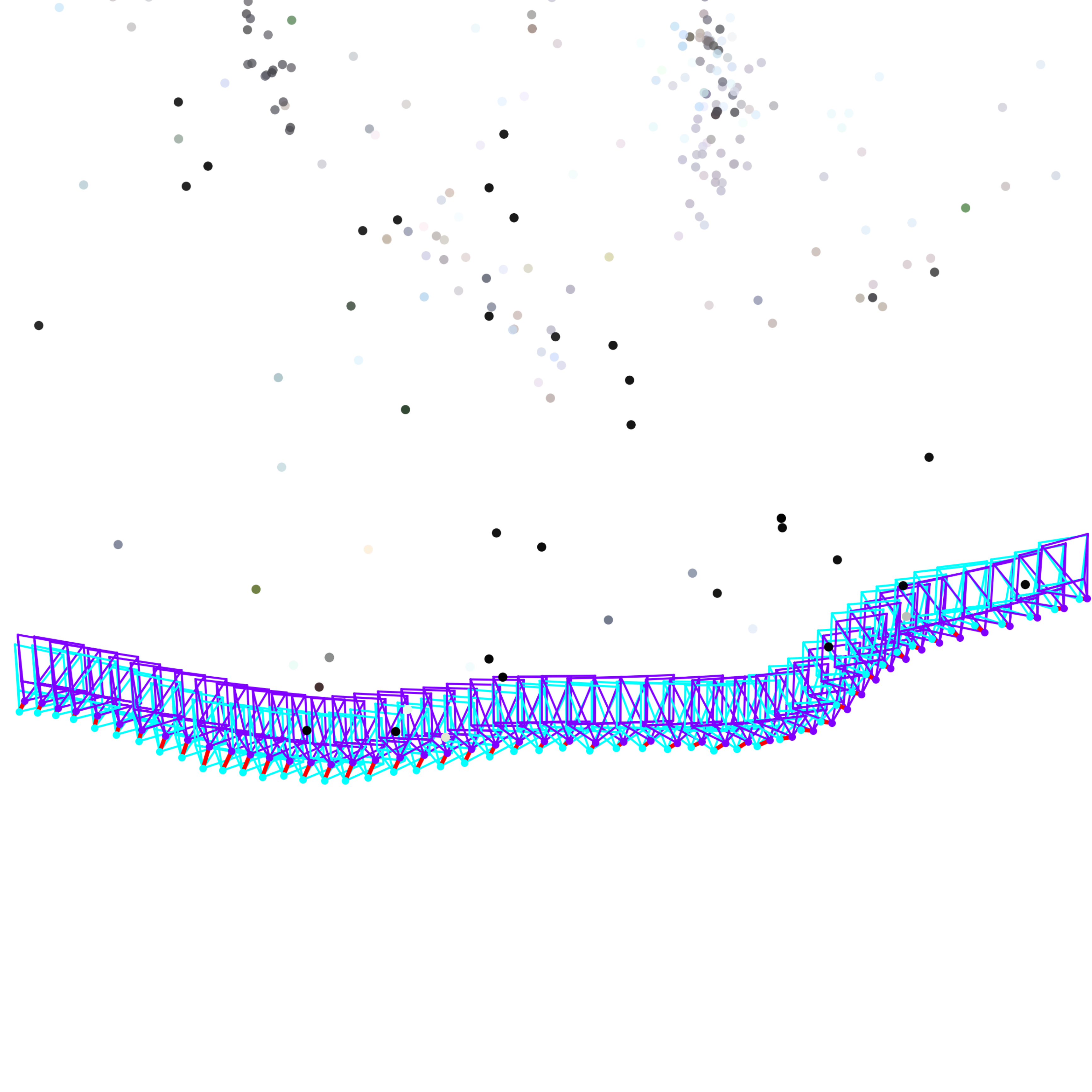}}

    \subfloat[RGB]{\includegraphics[height=0.18\linewidth, angle=-90]{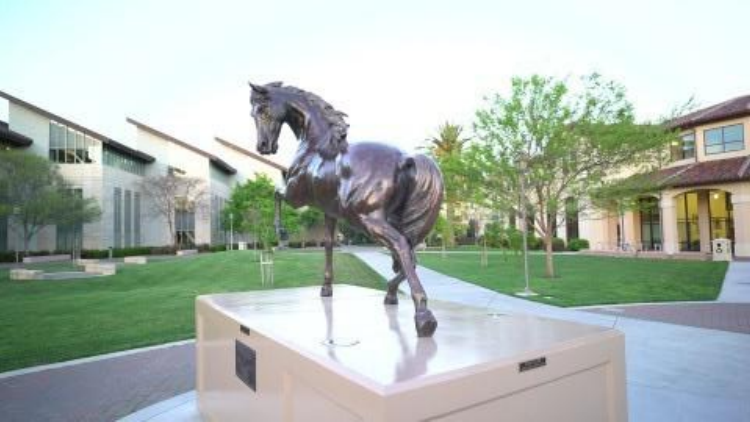}}
    \hspace{0.1em}
    \subfloat[Nope-NeRF \cite{bian2022nope}]{\includegraphics[height=0.18\linewidth, angle=-90]{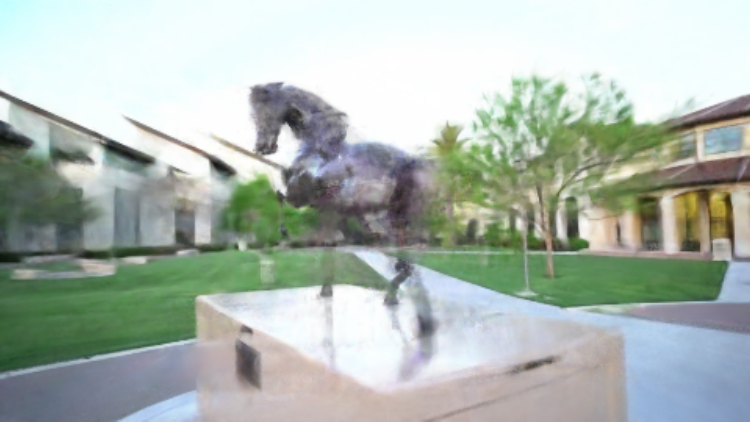}}
    \hspace{0.1em}
    \subfloat[CF-NeRF$^*$ \cite{yan2023cf}]{\includegraphics[height=0.18\linewidth, angle=-90]{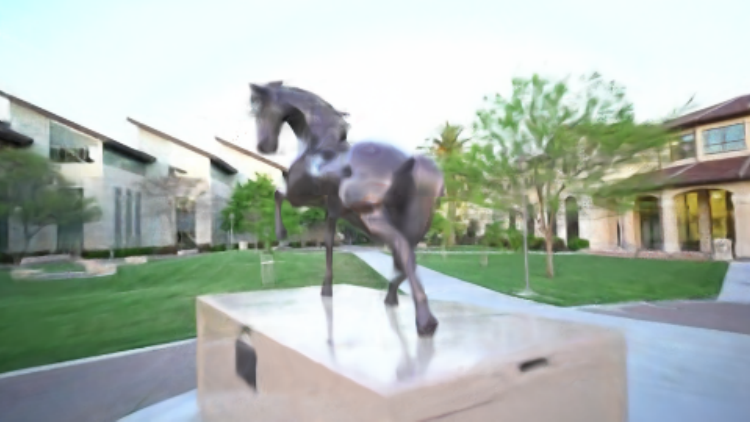}}
    \hspace{0.1em}
    \subfloat[CF-3DGS \cite{fu2024cf3dgs}]{\includegraphics[height=0.18\linewidth, angle=-90]{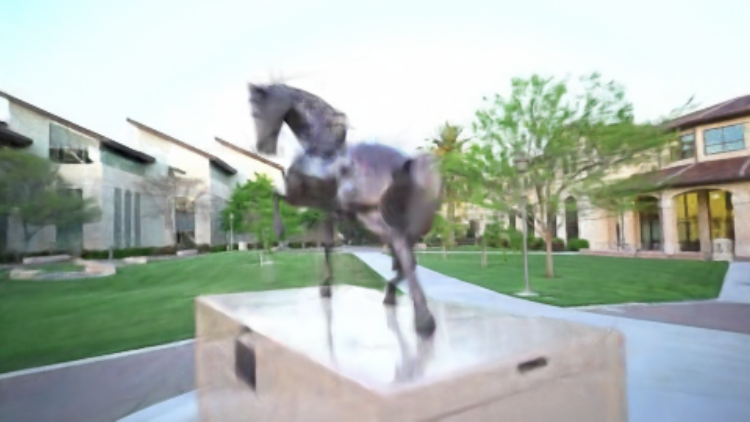}}
    \hspace{0.1em}
    \subfloat[RA-NeRF]{\includegraphics[height=0.18\linewidth, angle=-90]{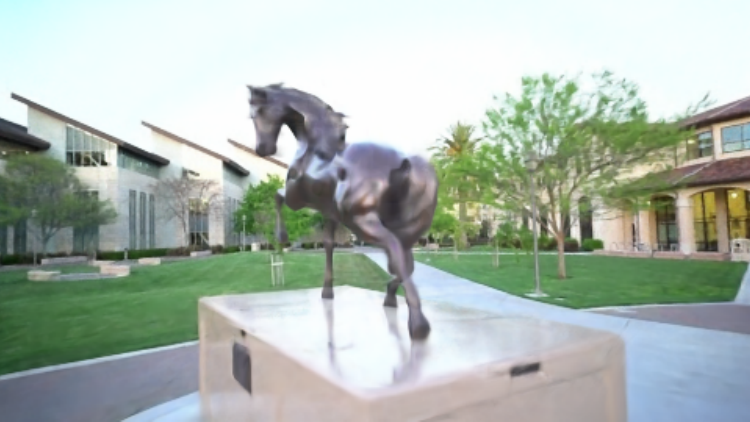}}

    \vspace{-0.5em}

    \caption{
        Visualizations of estimated camera poses and rendering results.
        (a)-(c) are results from NeRFBuster \cite{warburg2023nerfbusters} and (f)-(j) are results from Tanks\&Temple \cite{knapitsch2017tanks}.
        } 
    \label{fig:viz_poses_in_roses}

    \vspace{-2.0em}

\end{figure*}

\subsection{Incremental pipeline}
\label{sec:incremental_pipeline}

RA-NeRF uses the incremental pipeline to avoid converging to local minima by estimating camera poses progressively, which contains initialization, localization, partial optimization, and global optimization, with enhancements from flow-driven pose regulation and implicit pose filter.

\paragraph{Loss Function}

The loss function of RA-NeRF consists of two components, as shown in Eq. \ref{eq:loss}, where $\mathcal{L}_p$ represents photometric consistency, $\lambda_i$ and $ \lambda_f$ are weights for the flow-driven pose regulation.

\begin{equation}
    \label{eq:loss}
	\mathcal{L} = \mathcal{L}_p + \lambda_i \lambda_f \mathcal{L}_f 
\end{equation}

\paragraph{Initialization}

Initialization is the first and most critical step in camera pose estimation, forming the foundation for subsequent processing.
Instead of the complex strategies used in \cite{cheng2023lu,yan2023cf,ran2024ct}, RA-NeRF applies photometric consistency and flow-driven pose regulation during initialization.
After setting all camera poses to the identity, RA-NeRF begins by randomly initializing the implicit pose filter and the NeRF model.
Next, RA-NeRF selects the first $N_{init}=2$ images and estimates the camera pose using Eq. \ref{eq:loss} with $\xi_{iter}$ iterations, where $\lambda_i=1$.

\paragraph{Localization}

After initialization, RA-NeRF needs to determine the camera pose for a newly added image $I_{new}$.
As NeRF performs poorly from viewpoints with no observations, RA-NeRF leverages both photometric consistency and flow-driven pose regulation for accurate localization.
RA-NeRF first initializes the camera pose of $I_{new}$ by previous image, then selects $N_{loc}$ previous images $\mathcal{I}_{loc}$, reinitializes the implicit pose filter, reloads the NeRF model from the previous stage, computes the optical flow between $I_{new}$ and $\mathcal{I}_{loc}$, fixes camera poses of $\mathcal{I}_{loc}$, and finally estimates the camera pose of $I_{new}$ by Eq. \ref{eq:loss} with $\xi_{loc}$ iterations, where $\lambda_i=1$.
When $N_{loc} \geq 2$, the scale can be guaranteed to remain constant according to multi-view geometry \cite{hartley2003multiple}.

\paragraph{Optimization}

After localization, RA-NeRF performs partial optimization on $I_{new}$ and previous $N_{part}$ images with the NeRF model to minimize drift.
When every $N_{glob}$ new images are added, RA-NeRF performs global optimization on all images and the NeRF model.
During optimization, RA-NeRF does not utilize the flow-driven pose regulation, as it is only effective for neighboring images and does not apply well to a larger set of images.
Therefore, RA-NeRF uses Eq. \ref{eq:loss} with $\xi_{part}$ and $\xi_{glob}$ iterations, where $\lambda_i=0$. 
Besides, RA-NeRF reinitializes the implicit pose filter and reloads the NeRF model from the previous stage.

\section{Experiments}

\begin{table*}
    \begin{center}

        \vspace{-0.5em}

        \caption{ 
        Quantitative results on the NeRFBuster dataset, characterized by large rotational motions.
        RA-NeRF achieves state-of-the-art results, outperforming NeRFmm \cite{wang2021nerf}, SiRENmm \cite{ventusff2021}, BARF \cite{lin2021barf}, GARF \cite{chng2022garf}, L2G-NeRF \cite{chen2023local}, CF-NeRF \cite{yan2023cf}, Nope-NeRF \cite{bian2022nope}, and CF-3DGS \cite{fu2024cf3dgs}.
        The \colorbox{red!25}{\textbf{best}} results and the  \colorbox{yellow!25}{\underline{second best}} results are highlighted.
        }
        \label{tab:pose_results_nerfbuster}

        \vspace{-0.5em}
        
        \resizebox{\linewidth}{!}{\begin{tabular}{c|l|c|cccccccccccc}
                
                \hline\noalign{\smallskip}
                


                & & Mean & aloe & art & car & century & flowers & garbage 
                & picnic & pikachu & pipe & plant & roses & table \\

                \hline

                \multirow{10}{*}{$\Delta R \downarrow$}
                & NeRFmm & 79.1905
                & 149.0038 & 13.7975 & 62.8193 & 86.2631 & 93.8266 & 88.8254
                & 52.0380 & 87.3465 & 43.8570 & 27.7722 & 104.0850 & 104.6515 \\
                & SiRENmm & 81.2834
                & 135.3456 & 12.4257 & 87.8376 & 94.5317 & 92.2866 & 60.5024
                & 45.0568 & 26.7353 & 46.8526 & 133.9530 & 139.5724 & 100.3015 \\
                & BARF  & 77.6321
                & 154.1543 & 14.7550 & 81.3786 & 64.2474 & 74.8855 & 75.4210
                & 61.1011 & 33.5440 & 47.4349 & 88.8684 & 125.3059 & 110.4894 \\
                & GARF  & 80.3641
                & 118.9543 & 14.4842 & 56.7652 & 72.6237 & 97.8492 & 71.9970
                & 49.3072 & 34.3633 & 106.4186 & 101.3184 & 116.4942 & 123.7941 \\
                & L2G-NeRF & 66.0287
                & 65.7245 & 13.3651 & 31.8843 & 83.9665 & 78.3117 & 86.0642
                & 60.9882 & 97.6565 & 23.4746 & 37.6566 & 81.5597 & 131.6929 \\
                & Nope-NeRF & 56.1676
                & 48.7694 & 9.2974 & 68.3788 & 38.7692 & 57.5804 & 73.8243
                & 43.3736 & 23.8043 & 29.1478 & 154.0849 & 57.3135 & 69.6677 \\
                & CF-NeRF & 9.1090
                & \cellcolor{yellow!25}\underline{2.5324}
                & 18.7037 & 17.7240 
                & \cellcolor{yellow!25}\underline{6.1463}
                & 6.3273 & 8.5708
                & 10.8273 
                & \cellcolor{yellow!25}\underline{10.4663}
                & 16.7156 & 4.4945 & 2.7300 
                & \cellcolor{yellow!25}\underline{4.0700} \\
                & CF-NeRF$^*$ 
                & \cellcolor{yellow!25}\underline{7.0779}
                & 4.5319
                & 4.7976
                & 9.0527 & 11.9256 
                & \cellcolor{yellow!25}\underline{4.0435}
                & \cellcolor{yellow!25}\underline{3.2722}
                & 13.4793 & 15.7543 
                & \cellcolor{yellow!25}\underline{7.6741}
                & \cellcolor{yellow!25}\underline{3.1088}
                & \cellcolor{yellow!25}\underline{1.6278}
                & 5.6671 \\
                & CF-3DGS & 11.2151
                & 16.6094 
                & \cellcolor{yellow!25}\underline{3.8206} 
                & \cellcolor{yellow!25}\underline{4.5792}
                & 25.6687 & 12.8236 & 5.6919
                & \cellcolor{yellow!25}\underline{3.1399}
                & 16.1435 & 5.3298 & 8.6912 & 19.8090 & 12.2740 \\
                & RA-NeRF 
                & \cellcolor{red!25}\textbf{1.1449}
                & \cellcolor{red!25}\textbf{0.8144} 
                & \cellcolor{red!25}\textbf{0.3978} 
                & \cellcolor{red!25}\textbf{0.9128}  
                & \cellcolor{red!25}\textbf{1.1209} 
                & \cellcolor{red!25}\textbf{0.6981} 
                & \cellcolor{red!25}\textbf{0.9752}
                & \cellcolor{red!25}\textbf{1.4214} 
                & \cellcolor{red!25}\textbf{1.1002} 
                & \cellcolor{red!25}\textbf{2.6337} 
                & \cellcolor{red!25}\textbf{1.3986} 
                & \cellcolor{red!25}\textbf{0.5753} 
                & \cellcolor{red!25}\textbf{1.6909} \\
                
                \hline

                \multirow{10}{*}{$\Delta T \downarrow$}
                & NeRFmm  & 25.2413
                & 25.2792 & 22.1417 & 37.1552 & 35.4295 & 25.7556 & 21.0362
                & 16.8080 & 40.5621 & 9.4368 & 19.9647 & 24.1385 & 25.1881 \\
                & SiRENmm  & 25.4806
                & 27.7864 & 23.4831 & 42.9243 & 39.9758 & 26.8198 & 16.4988
                & 12.2677 & 5.3457 & 10.6975 & 46.4030 & 28.2898 & 25.2755 \\
                & BARF  & 22.9426
                & 19.7298 & 20.6491 & 42.3339 & 26.3996 & 20.6169 & 19.7850
                & 11.4757 & 21.6042 & 16.1818 & 35.0209 & 19.0742 & 22.4406 \\
                & GARF & 21.4076
                & 20.8661 & 22.9420 & 29.1316 & 28.2595 & 21.8617 & 18.2010
                & 14.4922 & 13.3614 & 17.4009 & 38.9866 & 10.7209 & 20.6676 \\
                & L2G-NeRF & 19.9316
                & 24.1720 & 13.1701 & 19.8383 & 26.8447 & 13.2949 & 16.8679
                & 13.9380 & 43.4267 & 13.6629 & 18.5277 & 19.0983 & 16.3371 \\
                & Nope-NeRF & 21.1305
                & 26.8229 & 18.2390 & 38.4676 & 11.5269 & 19.0489 & 33.5241
                & 10.3129 & 11.6039 & 13.8901 & 34.0780 & 20.3076 & 15.7444 \\
                & CF-NeRF  & 2.6209
                & 1.1089 & 8.5884 & 4.9247 
                & \cellcolor{yellow!25}\underline{1.4121}
                & 1.4058 & 3.4755
                & \cellcolor{yellow!25}\underline{0.7476} 
                & \cellcolor{yellow!25}\underline{0.9453}
                & 5.8665 & 1.1983 & 0.8495 
                & \cellcolor{yellow!25}\underline{0.9280} \\
                & CF-NeRF$^*$  & \cellcolor{yellow!25}\underline{1.8853}
                & \cellcolor{yellow!25}\underline{0.7474}
                & 3.4286 & 2.1918 & 2.0726 
                & \cellcolor{yellow!25}\underline{0.7475}
                & \cellcolor{yellow!25}\underline{2.3736}
                & 4.0586 & 1.7166 
                & \cellcolor{yellow!25}\underline{2.6637} 
                & \cellcolor{yellow!25}\underline{0.6550}
                & \cellcolor{yellow!25}\underline{0.2803}
                & 1.6879 \\
                & CF-3DGS & 3.6402
                & 3.7368 
                & \cellcolor{yellow!25}\underline{2.7370} 
                & \cellcolor{yellow!25}\underline{1.5525}
                & 6.6890 & 5.1875 & 5.9701
                & 1.3189 & 1.8616 & 3.1326 & 2.1379 & 6.0114 & 3.3472 \\
                & RA-NeRF 
                & \cellcolor{red!25}\textbf{0.5388}
                & \cellcolor{red!25}\textbf{0.5255} 
                & \cellcolor{red!25}\textbf{0.6629} 
                & \cellcolor{red!25}\textbf{0.3912}  
                & \cellcolor{red!25}\textbf{0.1876} 
                & \cellcolor{red!25}\textbf{0.4147} 
                & \cellcolor{red!25}\textbf{1.1159} 
                & \cellcolor{red!25}\textbf{0.4917} 
                & \cellcolor{red!25}\textbf{0.4424} 
                & \cellcolor{red!25}\textbf{0.9627} 
                & \cellcolor{red!25}\textbf{0.2715} 
                & \cellcolor{red!25}\textbf{0.1954} 
                & \cellcolor{red!25}\textbf{0.8046} \\
                
			\hline

                \multirow{10}{*}{PSNR $\uparrow$}
                & NeRFmm  & 16.2421
                & 19.5471 & 15.3702 & 14.9577 & 13.9959 & 16.6522 & 14.5127
                & 15.7169 & 20.5398 & 17.0115 & 18.0459 & 14.7628 & 13.7927 \\
                & SiRENmm  & 19.9123
                & 22.8658 & 18.3452 & 20.6571 & 17.3565 & 18.8590 & 17.6743
                & 20.6531 & 25.3286 & 20.5824 & 24.2302 & 15.6283 & 16.7668 \\
                & BARF    & 18.6288
                & 21.9318 & 20.2260 & 17.1700 & 15.2268 & 17.5742 & 16.6950
                & 19.3063 & 22.3343 & 19.8260 & 26.1715 & 13.5690 & 13.5144 \\
                & GARF   & 18.4584
                & 21.7836 & 19.9415 & 17.1403 & 15.6205 & 17.5725 & 17.0494
                & 19.2293 & 22.2954 & 17.8705 & 25.5554 & 13.9215 & 13.5205 \\
                & L2G-NeRF   & 17.9168
                & 20.6768 & 19.2531 & 17.7929 & 15.2495 & 18.1073 & 14.2474
                & 17.6790 & 21.3126 & 19.0025 & 19.5653 & 16.7540 & 15.3606 \\
                & Nope-NeRF   & 18.6676
                & 22.8376 & 19.6087 & 19.8081 & 16.2612 & 16.0533 & 15.5310
                & 18.9618 & 26.6462 & 18.7508 & 19.9893 & 15.4433 & 14.1197 \\
                & CF-NeRF    & 24.4286
                & 27.3675 
                & \cellcolor{yellow!25}\underline{26.7849}
                & 22.5143 & 21.4383 & 21.8342 
                & \cellcolor{yellow!25}\underline{24.3138}
                & 21.8609 
                & \cellcolor{yellow!25}\underline{31.9826}
                & \cellcolor{yellow!25}\underline{22.0737}
                & 25.6636 & 25.6908 & 21.6181 \\
                & CF-NeRF$^*$    
                & \cellcolor{yellow!25}\underline{25.3464}
                & \cellcolor{yellow!25}\underline{27.8416}
                & 25.2270 
                & \cellcolor{yellow!25}\underline{23.3041} 
                & \cellcolor{yellow!25}\underline{22.7553} 
                & \cellcolor{yellow!25}\underline{23.0537}
                & 24.0998
                & \cellcolor{yellow!25}\underline{23.1762}
                & 29.4807 & 21.9553 
                & \cellcolor{yellow!25}\underline{27.7628}
                & \cellcolor{yellow!25}\underline{27.0469} 
                & \cellcolor{yellow!25}\underline{28.4531} \\
                & CF-3DGS  & 22.0264
                & 24.7383 & 25.3289 & 20.0736 & 18.6802 & 17.4669 & 20.8899
                & 22.6554 & 30.3376 & 20.4041 & 24.1208 & 17.7880 & 21.8329 \\
                & RA-NeRF 
                & \cellcolor{red!25}\textbf{27.2538}
                & \cellcolor{red!25}\textbf{30.7039}
                & \cellcolor{red!25}\textbf{27.3680}
                & \cellcolor{red!25}\textbf{24.7297}
                & \cellcolor{red!25}\textbf{23.9631}
                & \cellcolor{red!25}\textbf{25.0211}
                & \cellcolor{red!25}\textbf{24.9725}
                & \cellcolor{red!25}\textbf{24.6056}
                & \cellcolor{red!25}\textbf{35.5195}
                & \cellcolor{red!25}\textbf{23.2117}
                & \cellcolor{red!25}\textbf{30.2112}
                & \cellcolor{red!25}\textbf{28.2220}
                & \cellcolor{red!25}\textbf{28.5171} \\

                \hline

                \multirow{10}{*}{LPIPS $\downarrow$}
                & NeRFmm  & 0.5719
                & 0.5326 & 0.4828 & 0.6275 & 0.6166 & 0.5955 & 0.5938
                & 0.6013 & 0.4500 & 0.5588 & 0.5676 & 0.5939 & 0.6426 \\
                & SiRENmm  & 0.4546
                & 0.4124 & 0.3853 & 0.4445 & 0.4883 & 0.5000 & 0.5331
                & 0.4962 & 0.3295 & 0.4872 & 0.3406 & 0.5416 & 0.4961 \\
                & BARF    & 0.4834
                & 0.3229 & 0.3451 & 0.5336 & 0.5723 & 0.5704 & 0.5349
                & 0.5213 & 0.3642 & 0.5039 & 0.3099 & 0.6203 & 0.6023 \\
                & GARF   & 0.4879
                & 0.3255 & 0.3595 & 0.5278 & 0.5459 & 0.5705 & 0.5398
                & 0.5269 & 0.3615 & 0.5608 & 0.3145 & 0.6231 & 0.5985 \\
                & L2G-NeRF   & 0.4920
                & 0.4593 & 0.4249 & 0.5349 & 0.5434 & 0.5467 & 0.5670
                & 0.5556 & 0.4306 & 0.4538 & 0.3851 & 0.4859 & 0.5169 \\
                & Nope-NeRF   & 0.4927
                & 0.4325 & 0.3595 & 0.4957 & 0.5188 & 0.6165 & 0.5601
                & 0.5245 & 0.3536 & 0.5475 & 0.4287 & 0.5337 & 0.5414 \\
                & CF-NeRF    & 0.3233
                & 0.1838 
                & \cellcolor{yellow!25}\underline{0.2256}
                & 0.4027 & 0.3608 & 0.3973 & 0.3774
                & \cellcolor{yellow!25}\underline{0.4750}
                & \cellcolor{yellow!25}\underline{0.1642}
                & 0.4573 & 0.2607 & 0.2767 & 0.2976 \\
                & CF-NeRF$^*$    
                & \cellcolor{yellow!25}\underline{0.3029}
                & \cellcolor{yellow!25}\underline{0.1705}
                & 0.2614 
                & \cellcolor{yellow!25}\underline{0.3788}
                & \cellcolor{yellow!25}\underline{0.3340}
                & \cellcolor{yellow!25}\underline{0.3424} 
                & \cellcolor{yellow!25}\underline{0.3444}
                & 0.4789 & 0.2142 
                & \cellcolor{yellow!25}\underline{0.4456} 
                & \cellcolor{yellow!25}\underline{0.1774} 
                & \cellcolor{yellow!25}\underline{0.2305} 
                & \cellcolor{yellow!25}\underline{0.2567} \\
                & CF-3DGS  & 0.4130
                & 0.3161 & 0.2817 & 0.4634 & 0.4886 & 0.5651 & 0.4201
                & 0.4691 & 0.2099 & 0.4913 & 0.3512 & 0.4996 & 0.3993 \\
                & RA-NeRF 
                & \cellcolor{red!25}\textbf{0.2452}
                & \cellcolor{red!25}\textbf{0.1067}
                & \cellcolor{red!25}\textbf{0.2175} 
                & \cellcolor{red!25}\textbf{0.3261}
                & \cellcolor{red!25}\textbf{0.2877}
                & \cellcolor{red!25}\textbf{0.2668}
                & \cellcolor{red!25}\textbf{0.2908}
                & \cellcolor{red!25}\textbf{0.4131} 
                & \cellcolor{red!25}\textbf{0.0907} 
                & \cellcolor{red!25}\textbf{0.3911} 
                & \cellcolor{red!25}\textbf{0.1286} 
                & \cellcolor{red!25}\textbf{0.1994} 
                & \cellcolor{red!25}\textbf{0.2234} \\

                \hline

		\end{tabular}}

        \vspace{-1.5em}

    \end{center}
\end{table*}

\begin{table*}
    \begin{center}

        \caption{ 
        Quantitative results on the Tanks\&Temple dataset.
        Compared with Nope-NeRF \cite{bian2022nope}, CF-NeRF \cite{yan2023cf}, and CF-3DGS \cite{fu2024cf3dgs}, RA-NeRF demonstrates competitive performance.
        The \colorbox{red!25}{\textbf{best}} results and the  \colorbox{yellow!25}{\underline{second best}} results are highlighted.
        }
        \label{tab:pose_results_tankstemple}

        \vspace{-0.5em}
        
        \resizebox{0.9\linewidth}{!}{\begin{tabular}{c|l|c|cccccccc}
                
                \hline\noalign{\smallskip}
                


                & & Mean & Ballroom & Barn & Church & Family 
                & Francis & Horse & Ignatius & Museum\\

                \hline

                \multirow{4}{*}{$\Delta R \downarrow$}
                & Nope-NeRF   
                & 58.4816
                & 2.0547 
                & 86.9369 
                & 145.0743 
                & 36.5905
                & \cellcolor{red!25}\textbf{1.0759} 
                & 163.9675 
                & \cellcolor{red!25}\textbf{0.5542} 
                & 31.5989 \\
                & CF-NeRF$^*$ 
                & 42.9849
                & \cellcolor{yellow!25}\underline{1.6731} 
                & 2.4796 
                & \cellcolor{yellow!25}\underline{4.1735} 
                & 171.8357
                & 149.7760 
                & \cellcolor{yellow!25}\underline{8.1537} 
                & 4.3137 
                & \cellcolor{red!25}\textbf{1.4737} \\
                & CF-3DGS     
                & \cellcolor{yellow!25}\underline{7.3920}
                & 5.8610 
                & \cellcolor{red!25}\textbf{1.2757} 
                & 16.4354 
                & \cellcolor{yellow!25}\underline{1.4195}
                & 2.8036 
                & 15.1154 
                & 9.4398 
                & 6.7858 \\
                & RA-NeRF     
                & \cellcolor{red!25}\textbf{1.9624}
                & \cellcolor{red!25}\textbf{0.5715} 
                & \cellcolor{yellow!25}\underline{3.2061} 
                & \cellcolor{red!25}\textbf{1.3643} 
                & \cellcolor{red!25}\textbf{0.8255}
                & \cellcolor{yellow!25}\underline{1.2063} 
                & \cellcolor{red!25}\textbf{4.9746} 
                & \cellcolor{yellow!25}\underline{1.5659} 
                & \cellcolor{yellow!25}\underline{1.9846} \\
                
                \hline

                \multirow{4}{*}{$\Delta T \downarrow$}
                & Nope-NeRF   
                & 5.3545
                & \cellcolor{yellow!25}\underline{0.3392} 
                & 5.0315 
                & 10.9355 
                & 5.6915
                & \cellcolor{yellow!25}\underline{0.4558} 
                & 1.1147 
                & \cellcolor{yellow!25}\underline{1.0528} 
                & 18.2152 \\
                & CF-NeRF$^*$ 
                & \cellcolor{yellow!25}\underline{1.3925}
                & 0.3562 
                & 0.5176 
                & \cellcolor{yellow!25}\underline{1.4012} 
                & 2.1876
                & 3.5242 
                & \cellcolor{yellow!25}\underline{0.5243} 
                & 2.3502 
                & \cellcolor{red!25}\textbf{0.2789} \\
                & CF-3DGS     
                & 2.2121
                & 3.4263 
                & \cellcolor{red!25}\textbf{0.3804} 
                & 5.0196 
                & \cellcolor{yellow!25}\underline{0.5425}
                & 0.9336 
                & 2.0865 & 
                2.7147 & 
                2.5930 \\
                & RA-NeRF     
                & \cellcolor{red!25}\textbf{0.3910}
                & \cellcolor{red!25}\textbf{0.2089} 
                & \cellcolor{yellow!25}\underline{0.4641} 
                & \cellcolor{red!25}\textbf{0.4469} 
                & \cellcolor{red!25}\textbf{0.1833}
                & \cellcolor{red!25}\textbf{0.2768} 
                & \cellcolor{red!25}\textbf{0.1767} 
                & \cellcolor{red!25}\textbf{0.8300} 
                & \cellcolor{yellow!25}\underline{0.5463} \\
                
			\hline

                \multirow{4}{*}{PSNR $\uparrow$}
                & Nope-NeRF  & 27.3324
                & \cellcolor{red!25}\textbf{31.4314}
                & 24.5962 & 23.6906 & 25.9189
                & 31.1863 & 24.2281 
                & \cellcolor{red!25}\textbf{29.3578} & 28.2500 \\
                & CF-NeRF$^*$ & 26.5533
                & 26.3096 
                & \cellcolor{yellow!25}\underline{25.6028} 
                & 27.2631 & 24.8720
                & 26.9767 & 26.3777 & 25.9892 
                & \cellcolor{yellow!25}\underline{29.0353} \\
                & CF-3DGS  
                & \cellcolor{yellow!25}\underline{27.6146}
                & 26.3057 & 23.8152 
                & \cellcolor{yellow!25}\underline{27.5475}
                & \cellcolor{yellow!25}\underline{30.4484}
                & \cellcolor{yellow!25}\underline{30.6915}
                & \cellcolor{yellow!25}\underline{26.6287} 
                & 26.9313 & 28.5483 \\
                & RA-NeRF 
                & \cellcolor{red!25}\textbf{29.8012}
                & \cellcolor{yellow!25}\underline{30.2495} 
                & \cellcolor{red!25}\textbf{25.9803} 
                & \cellcolor{red!25}\textbf{32.0474} 
                & \cellcolor{red!25}\textbf{30.4555}
                & \cellcolor{red!25}\textbf{32.0424}
                & \cellcolor{red!25}\textbf{28.2649}
                & \cellcolor{yellow!25}\underline{27.7772} 
                & \cellcolor{red!25}\textbf{31.5923} \\
                
			\hline

                \multirow{4}{*}{LPIPS $\downarrow$}
                & Nope-NeRF  & 0.2528
                & \cellcolor{red!25}\textbf{0.0892}
                & 0.4402 & 0.2429 & 0.3085
                & \cellcolor{red!25}\textbf{0.2303}
                & 0.3084 
                & \cellcolor{red!25}\textbf{0.2083} 
                & 0.1943 \\
                & CF-NeRF$^*$ & 0.2628
                & 0.1900 
                & \cellcolor{yellow!25}\underline{0.3849} 
                & \cellcolor{yellow!25}\underline{0.1353} 
                & 0.3225
                & 0.3851 
                & \cellcolor{yellow!25}\underline{0.2452} 
                & 0.2978 
                & \cellcolor{yellow!25}\underline{0.1415} \\
                & CF-3DGS  
                & \cellcolor{yellow!25}\underline{0.2340}
                & 0.2109 & 0.4076 & 0.1620 
                & \cellcolor{red!25}\textbf{0.1407}
                & 0.2674 & 0.2461 & 0.2592 & 0.1781 \\
                & RA-NeRF 
                & \cellcolor{red!25}\textbf{0.1844}
                & \cellcolor{yellow!25}\underline{0.1077}
                & \cellcolor{red!25}\textbf{0.3792} 
                & \cellcolor{red!25}\textbf{0.0773}
                & \cellcolor{yellow!25}\underline{0.1459}
                & \cellcolor{yellow!25}\underline{0.2350} 
                & \cellcolor{red!25}\textbf{0.1822} 
                & \cellcolor{yellow!25}\underline{0.2521} 
                & \cellcolor{red!25}\textbf{0.0958} \\

                \hline

		\end{tabular}}

        \vspace{-2.0em}

    \end{center}
\end{table*}

\subsection{Dataset}

To thoroughly evaluate the performance of various pose estimation methods, we use two datasets: NeRFBuster \cite{warburg2023nerfbusters,yan2023cf} and Tanks\&Temple \cite{knapitsch2017tanks,bian2022nope}.
The NeRFBuster dataset consists of 12 scenes, where most of scenes involve rotational motion around a central object and each scene includes approximately 50 images with a resolution of $480 \times 270$.
The Tanks\&Temple dataset comprises 8 real-world outdoor scenes, featuring a moving camera focused on a stationary object, where we uniformly sample around 50 images from each scene and resize them to a resolution of $480 \times 270$.

\subsection{Implementation}

We implement RA-NeRF using Pytorch with the Adam \cite{kingma2014adam}. 
On the NeRFBuster, we set the initial learning rate to $5e^{-4}$ and set $\lambda_f=1e^{-1}$. 
For the Tanks\&Temple, we initialize the learning rate to $5e^{-4}$, and set $\lambda_f=1e^{-3}$. 
Regarding other hyper-parameters, we set $\xi_{loca} = 1100$, $\xi_{init} = \xi_{part} = \xi_{glob} = 3100$, $N_{init}=N_{loc}=2$, $N_{part}=N_{glob}=5$, and $\lambda_r=3$.
All experiments are conducted on an NVIDIA RTX 3090 GPU.

\subsection{Evaluation}

We evaluate camera poses of different methods using rotation error $\Delta R$ and translation error $\Delta T$.
To calculate $\Delta R$ and $\Delta T$, we first align estimated camera centers with ground truth camera centers via a similarity transformation.
Then, we calculate the average Euler distance between translation vectors, and the average geodesic distance on the 3D manifold of rotation matrices.
Note that as we try to estimate camera poses for all images from scratch, we calculate $\Delta R$ and $\Delta T$ for all images.
To further compare the quality of camera poses and avoid the influence of different network structures, we uniformly use the NeRFacc \cite{li2022nerfacc} to evaluate novel view synthesis, where PSNR and LPIPS are calculated, where we sample one image for testing every eight images and use the remaining images for training.

\subsection{Results}

\paragraph{NeRFBuster}

Tab. \ref{tab:pose_results_nerfbuster} presents the quantitative results on the NeRFBuster dataset. 
Our method, RA-NeRF, achieves state-of-the-art performance across all metrics. 
NeRFmm \cite{wang2021nerf}, SiRENmm \cite{ventusff2021}, BARF \cite{lin2021barf}, GARF \cite{chng2022garf}, L2G-NeRF \cite{chen2023local}, and Nope-NeRF \cite{bian2022nope} produce incorrect camera poses, with $\Delta R$ exceeding $50^{\circ}$. 
These methods estimate camera poses for all images at once, leading to convergence to local minima when large rotation exists.
Although CF-NeRF \cite{yan2023cf} and CF-3DGS \cite{fu2024cf3dgs} use the incremental pipeline, they do not obtain high-quality camera poses. 
CF-NeRF solely relies on photometric consistency, facing drifts with more images, and CF-3DGS rapidly converges on representation, falling into local minima.
Fig. \ref{fig:viz_poses_in_roses} visualizes the results of different methods for better comparison.

\paragraph{Tanks\&Temple}

Tab. \ref{tab:pose_results_tankstemple} presents the results on the Tanks\&Temple dataset. 
Unlike the results on NeRFBuster, the differences between methods are smaller in this case, as this dataset primarily involves translations between images, instead of rotations, making it easier for methods to estimate camera poses.
However, it is still notable that CF-NeRF \cite{yan2023cf} and Nope-NeRF \cite{bian2022nope} fail on some scenes, as they converge to incorrect solutions highlighted by LU-NeRF \cite{cheng2023lu}.
RA-NeRF achieves state-of-the-art on most metrics, while CF-3DGS \cite{fu2024cf3dgs} ranks second.
Fig. \ref{fig:viz_poses_in_roses} visualizes the results of different methods for better comparison.

\subsection{Ablation}

To demonstrate the effectiveness of the proposed method, we conduct ablation experiments.
For clarity, we only report the mean values of the $\Delta R$ and $\Delta T$. 
Throughout ablation experiments, we utilize the NeRFBuster dataset.

\paragraph{Flow-driven pose regulation}

In the absence of an initial pose, photometric consistency by itself does not provide accurate gradient information and often leads to convergence to a local minimum.
However, with the flow-driven pose regulation, RA-NeRF can obtain better initial camera poses. 
Tab. \ref{tab:flow_constrain} presents the influences of the flow-driven pose regulation.
Without flow-driven pose regulation, RA-NeRF struggles to converge on several scenes, yielding camera poses with significant errors. 
However, by integrating flow-driven pose regulation during the initialization and localization, RA-NeRF significantly enhances the precision and robustness of pose estimation.

\begin{table}[tp]
    \begin{center}

        \caption{ 
        Impact of flow-driven pose regulation. 
        $\checkmark$ and $\times$ indicate the use and non-use of flow-driven pose regulation.
        }
        \label{tab:flow_constrain}

        \vspace{-0.5em}
        
        {\begin{tabular}{c|c|c|c}
                
            \hline\noalign{\smallskip}
            

            & Method & flow-driven pose regulation & Mean \\

            \hline

            \multirow{2}{*}{$\Delta R \downarrow$}
            & \multirow{2}{*}{RA-NeRF}
            & $\times$ & 78.0355 \\
            & & $\checkmark$ & \textbf{1.1449} \\
            
            \hline

            \multirow{2}{*}{$\Delta T \downarrow$}
            & \multirow{2}{*}{RA-NeRF}
            & $\times$ & 20.9749 \\
            & & $\checkmark$ & \textbf{0.5388} \\
                
			\hline

		\end{tabular}}

        \vspace{-1.5em}

    \end{center}
\end{table}

\begin{table}[tp]
    \begin{center}

        \caption{ 
        Impact of implicit pose filter.
        $\checkmark$ and $\times$ indicate the use and non-use of the corresponding module, where direct refers to directly calculating gradients on $\mathfrak{se}(3)$ vector.
        }
        \label{tab:implicit_pose}

        \vspace{-0.5em}

        {\begin{tabular}{c|c|ccc|c}
                
            \hline\noalign{\smallskip}

            & \multirow{2}{*}{Method} 
            & \multirow{2}{*}{direct} 
            & \multicolumn{2}{c|}{implicit}
            & \multirow{2}{*}{Mean} \\
            \cline{4-5}
            & & & local & global & \\

            \hline

            \multirow{4}{*}{$\Delta R \downarrow $}
            & \multirow{4}{*}{RA-NeRF}
            & $\checkmark$ & $\times$ & $\times$ & 8.8946 \\
            & & $\times$ & $\checkmark$ & $\times$ & 1.5056 \\
            & & $\checkmark$ & $\checkmark$ & $\checkmark$ & 1.1889 \\
            & & $\times$ & $\checkmark$ & $\checkmark$ & \textbf{1.1449} \\
            
            \hline

            \multirow{4}{*}{$\Delta T \downarrow $}
            & \multirow{4}{*}{RA-NeRF}
            & $\checkmark$ & $\times$ & $\times$ & 2.3920 \\
            & & $\times$ & $\checkmark$ & $\times$ & 0.6179 \\
            & & $\checkmark$ & $\checkmark$ & $\checkmark$ & 0.5815 \\
            & & $\times$ & $\checkmark$ & $\checkmark$ & \textbf{0.5388} \\
                
			\hline

		\end{tabular}}

        \vspace{-3.0em}

    \end{center}
\end{table}

\begin{table}[tp]
    \begin{center}

        \caption{ 
        Impact of pose updating strategies.
        }
        \label{tab:estimate_mode}

        \vspace{-0.5em}

        {\begin{tabular}{c|c|c|c}
                
            \hline\noalign{\smallskip}

            & Method & pose update domain & Mean \\

            \hline

            \multirow{2}{*}{$\Delta R \downarrow$}
            & \multirow{2}{*}{RA-NeRF}
            & $\mathfrak{se}(3)$ & 5.5228 \\
            & & SE(3) & \textbf{1.1449} \\

            \hline

            \multirow{2}{*}{$\Delta T \downarrow $}
            & \multirow{2}{*}{RA-NeRF}
            & $\mathfrak{se}(3)$ & 3.2336 \\
            & & SE(3) & \textbf{0.5388} \\
                
			\bottomrule

		\end{tabular}}

        \vspace{-3.0em}

    \end{center}
\end{table}

\paragraph{Implicit pose filter}

To deal with noise from photometric consistency and flow-driven pose regulation, IF-NeRF proposes the implicit pose filter with local motion embedding and global motion embedding to update camera poses.
In contrast, the straightforward solution is directly setting a six-dimensional $\mathfrak{se}(3)$ vector as learnable parameters to estimate the pose.
Tab. \ref{tab:implicit_pose} compares the influences of different methods.
RA-NeRF gives the worst results when the $\mathfrak{se}(3)$ vector is used directly.
Even when we combine the direct and implicit methods, the results of RA-NeRF are still worse than the implicit method.
In addition, we note that without using global motion embedding, RA-NeRF is unable to model the motion of the whole scene and still degrades the accuracy.

Meanwhile, we also investigate different strategies to update camera poses. 
In addition to updating the camera pose in SE(3), another approach is to convert $P_{init}$ to $\mathfrak{se}(3)$, then update it by adding $\Delta \mathfrak{se}3$, and finally convert it back to SE(3). 
Tab. \ref{tab:estimate_mode} shows that updating in $\mathfrak{se}(3)$ yields worse results. 
Although SE(3) and $\Delta \mathfrak{se}3$ are equivalent, there are multiple data transformations in $\Delta \mathfrak{se}3$, raising the risk of numerical computational instability.

\subsection{Limitation}

Although RA-NeRF achieves state-of-the-art results in pose estimation, there are several limitations.
On the one hand, RA-NeRF requires overlapping images to build pixel correspondences.
On the other hand, RA-NeRF is built upon NeRF with high computation complexity.
However, these limitations can be overcome by a robust matching method, such as Dust3R \cite{wang2024dust3r}, and an efficient backbone, such as Instant-NGP \cite{muller2022instant} or 3DGS \cite{kerbl3Dgaussians}.
\section{Conclusion}

In this paper, we propose RA-NeRF, a novel approach to estimate high quality camera poses from complex trajectories.
Following the incremental pipeline, RA-NeRF combines photometric consistency and flow-driven pose regulation to avoid converging to local minima, and leverages the implicit pose filter to deal with noise in gradients and complex motions.
Experiments on the NeRFBuster and Tanks\&Temple demonstrate that RA-NeRF outperforms existing methods and achieves state-of-the-art results.
In summary, RA-NeRF demonstrates that it is possible to obtain high-quality camera poses and presents a promising end-to-end pipeline to reconstruct the 3D scene from unposed images.

\bibliographystyle{IEEEtran}
\bibliography{IEEEexample}

\begin{thebibliography}{10}
\providecommand{\url}[1]{#1}
\csname url@rmstyle\endcsname
\providecommand{\newblock}{\relax}
\providecommand{\bibinfo}[2]{#2}
\providecommand\BIBentrySTDinterwordspacing{\spaceskip=0pt\relax}
\providecommand\BIBentryALTinterwordstretchfactor{4}
\providecommand\BIBentryALTinterwordspacing{\spaceskip=\fontdimen2\font plus
\BIBentryALTinterwordstretchfactor\fontdimen3\font minus
  \fontdimen4\font\relax}
\providecommand\BIBforeignlanguage[2]{{%
\expandafter\ifx\csname l@#1\endcsname\relax
\typeout{** WARNING: IEEEtran.bst: No hyphenation pattern has been}%
\typeout{** loaded for the language `#1'. Using the pattern for}%
\typeout{** the default language instead.}%
\else
\language=\csname l@#1\endcsname
\fi
#2}}

\bibitem{bian2022nope}
W.~Bian, Z.~Wang, K.~Li, J.-W. Bian, and V.~A. Prisacariu, ``Nope-nerf:
  Optimising neural radiance field with no pose prior,'' in \emph{Proceedings
  of the IEEE/CVF Conference on Computer Vision and Pattern Recognition}, 2023,
  pp. 4160--4169.

\bibitem{yan2023cf}
Q.~Yan, Q.~Wang, K.~Zhao, J.~Chen, B.~Li, X.~Chu, and F.~Deng, ``Cf-nerf:
  Camera parameter free neural radiance fields with incremental learning,'' in
  \emph{Proceedings of the AAAI Conference on Artificial Intelligence},
  vol.~38, no.~6, 2024, pp. 6440--6448.

\bibitem{fu2024cf3dgs}
Y.~Fu, S.~Liu, A.~Kulkarni, J.~Kautz, A.~A. Efros, and X.~Wang, ``Colmap-free
  3d gaussian splatting,'' in \emph{Proceedings of the IEEE/CVF Conference on
  Computer Vision and Pattern Recognition (CVPR)}, June 2024, pp.
  20\,796--20\,805.

\bibitem{warburg2023nerfbusters}
F.~Warburg, E.~Weber, M.~Tancik, A.~Holynski, and A.~Kanazawa, ``Nerfbusters:
  Removing ghostly artifacts from casually captured nerfs,'' \emph{arXiv
  preprint arXiv:2304.10532}, 2023.

\bibitem{mildenhall2021nerf}
B.~Mildenhall, P.~P. Srinivasan, M.~Tancik, J.~T. Barron, R.~Ramamoorthi, and
  R.~Ng, ``Nerf: Representing scenes as neural radiance fields for view
  synthesis,'' in \emph{Computer Vision--ECCV 2020: 16th European Conference,
  Glasgow, UK, August 23--28, 2020, Proceedings, Part I}, 2020, pp. 405--421.

\bibitem{kerbl3Dgaussians}
B.~Kerbl, G.~Kopanas, T.~Leimk{\"u}hler, and G.~Drettakis, ``3d gaussian
  splatting for real-time radiance field rendering,'' \emph{ACM Transactions on
  Graphics}, vol.~42, no.~4, July 2023.

\bibitem{pumarola2021d}
A.~Pumarola, E.~Corona, G.~Pons-Moll, and F.~Moreno-Noguer, ``D-nerf: Neural
  radiance fields for dynamic scenes,'' in \emph{Proceedings of the IEEE/CVF
  Conference on Computer Vision and Pattern Recognition}, 2021, pp.
  10\,318--10\,327.

\bibitem{li2023dynibar}
Z.~Li, Q.~Wang, F.~Cole, R.~Tucker, and N.~Snavely, ``Dynibar: Neural dynamic
  image-based rendering,'' in \emph{Proceedings of the IEEE/CVF Conference on
  Computer Vision and Pattern Recognition}, 2023.

\bibitem{li2023neuralangelo}
Z.~Li, T.~M{\"u}ller, A.~Evans, R.~H. Taylor, M.~Unberath, M.-Y. Liu, and C.-H.
  Lin, ``Neuralangelo: High-fidelity neural surface reconstruction,'' in
  \emph{Proceedings of the IEEE/CVF Conference on Computer Vision and Pattern
  Recognition}, 2023, pp. 8456--8465.

\bibitem{chen2024pgsr}
D.~Chen, H.~Li, W.~Ye, Y.~Wang, W.~Xie, S.~Zhai, N.~Wang, H.~Liu, H.~Bao, and
  G.~Zhang, ``Pgsr: Planar-based gaussian splatting for efficient and
  high-fidelity surface reconstruction,'' \emph{arXiv preprint
  arXiv:2406.06521}, 2024.

\bibitem{zhenxingswitch}
M.~Zhenxing and D.~Xu, ``Switch-nerf: Learning scene decomposition with mixture
  of experts for large-scale neural radiance fields,'' in \emph{International
  Conference on Learning Representations}, 2023.

\bibitem{lin2024vastgaussian}
J.~Lin, Z.~Li, X.~Tang, J.~Liu, S.~Liu, J.~Liu, Y.~Lu, X.~Wu, S.~Xu, Y.~Yan,
  \emph{et~al.}, ``Vastgaussian: Vast 3d gaussians for large scene
  reconstruction,'' in \emph{Proceedings of the IEEE/CVF Conference on Computer
  Vision and Pattern Recognition}, 2024, pp. 5166--5175.

\bibitem{schonberger2016structure}
J.~L. Schonberger and J.-M. Frahm, ``Structure-from-motion revisited,'' in
  \emph{Proceedings of the IEEE conference on computer vision and pattern
  recognition}, 2016, pp. 4104--4113.

\bibitem{wang2021nerf}
Z.~Wang, S.~Wu, W.~Xie, M.~Chen, and V.~A. Prisacariu, ``Nerf--: Neural
  radiance fields without known camera parameters,'' \emph{arXiv preprint
  arXiv:2102.07064}, 2021.

\bibitem{lin2021barf}
C.-H. Lin, W.-C. Ma, A.~Torralba, and S.~Lucey, ``Barf: Bundle-adjusting neural
  radiance fields,'' in \emph{Proceedings of the IEEE/CVF International
  Conference on Computer Vision}, 2021, pp. 5741--5751.

\bibitem{ventusff2021}
J.~Guo and A.~Sherwood, ``imporved-nerfmm,''
  \url{https://github.com/ventusff/improved-nerfmm}, 2021.

\bibitem{chng2022garf}
S.-F. Chng, S.~Ramasinghe, J.~Sherrah, and S.~Lucey, ``Gaussian activated
  neural radiance fields for high fidelity reconstruction and pose
  estimation,'' in \emph{European Conference on Computer Vision}.\hskip 1em
  plus 0.5em minus 0.4em\relax Springer, 2022, pp. 264--280.

\bibitem{chen2023local}
Y.~Chen, X.~Chen, X.~Wang, Q.~Zhang, Y.~Guo, Y.~Shan, and F.~Wang,
  ``Local-to-global registration for bundle-adjusting neural radiance fields,''
  in \emph{Proceedings of the IEEE/CVF Conference on Computer Vision and
  Pattern Recognition}, 2023, pp. 8264--8273.

\bibitem{cheng2023lu}
Z.~Cheng, C.~Esteves, V.~Jampani, A.~Kar, S.~Maji, and A.~Makadia, ``Lu-nerf:
  Scene and pose estimation by synchronizing local unposed nerfs,'' in
  \emph{Proceedings of the IEEE/CVF International Conference on Computer
  Vision}, 2023, pp. 18\,312--18\,321.

\bibitem{mildenhall2019llff}
B.~Mildenhall, P.~P. Srinivasan, R.~Ortiz-Cayon, N.~K. Kalantari,
  R.~Ramamoorthi, R.~Ng, and A.~Kar, ``Local light field fusion: Practical view
  synthesis with prescriptive sampling guidelines,'' \emph{ACM Transactions on
  Graphics (TOG)}, 2019.

\bibitem{tan2024td}
Z.~Tan, Z.~Zhou, Y.~Ge, Z.~Wang, X.~Chen, and D.~Hu, ``Td-nerf: Novel truncated
  depth prior for joint camera pose and neural radiance field optimization,''
  in \emph{2024 IEEE/RSJ International Conference on Intelligent Robots and
  Systems (IROS)}.\hskip 1em plus 0.5em minus 0.4em\relax IEEE, 2024, pp.
  372--379.

\bibitem{ranftl2021vision}
R.~Ranftl, A.~Bochkovskiy, and V.~Koltun, ``Vision transformers for dense
  prediction,'' in \emph{Proceedings of the IEEE/CVF international conference
  on computer vision}, 2021, pp. 12\,179--12\,188.

\bibitem{jeong2021self}
Y.~Jeong, S.~Ahn, C.~Choy, A.~Anandkumar, M.~Cho, and J.~Park,
  ``Self-calibrating neural radiance fields,'' in \emph{Proceedings of the
  IEEE/CVF International Conference on Computer Vision}, 2021, pp. 5846--5854.

\bibitem{bian2023porf}
J.-W. Bian, W.~Bian, V.~A. Prisacariu, and P.~Torr, ``Porf: Pose residual field
  for accurate neural surface reconstruction,'' \emph{arXiv preprint
  arXiv:2310.07449}, 2023.

\bibitem{truong2023sparf}
P.~Truong, M.-J. Rakotosaona, F.~Manhardt, and F.~Tombari, ``Sparf: Neural
  radiance fields from sparse and noisy poses,'' in \emph{Proceedings of the
  IEEE/CVF Conference on Computer Vision and Pattern Recognition}, 2023, pp.
  4190--4200.

\bibitem{truong2021learning}
P.~Truong, M.~Danelljan, L.~Van~Gool, and R.~Timofte, ``Learning accurate dense
  correspondences and when to trust them,'' in \emph{Proceedings of the
  IEEE/CVF conference on computer vision and pattern recognition}, 2021, pp.
  5714--5724.

\bibitem{sun2021loftr}
J.~Sun, Z.~Shen, Y.~Wang, H.~Bao, and X.~Zhou, ``Loftr: Detector-free local
  feature matching with transformers,'' in \emph{Proceedings of the IEEE/CVF
  conference on computer vision and pattern recognition}, 2021, pp. 8922--8931.

\bibitem{meng2021gnerf}
Q.~Meng, A.~Chen, H.~Luo, M.~Wu, H.~Su, L.~Xu, X.~He, and J.~Yu, ``Gnerf:
  Gan-based neural radiance field without posed camera,'' in \emph{Proceedings
  of the IEEE/CVF International Conference on Computer Vision}, 2021, pp.
  6351--6361.

\bibitem{zhang2023pose}
J.~Zhang, F.~Zhan, Y.~Yu, K.~Liu, R.~Wu, X.~Zhang, L.~Shao, and S.~Lu,
  ``Pose-free neural radiance fields via implicit pose regularization,'' in
  \emph{Proceedings of the IEEE/CVF International Conference on Computer
  Vision}, 2023, pp. 3534--3543.

\bibitem{chen2023dbarf}
Y.~Chen and G.~H. Lee, ``Dbarf: Deep bundle-adjusting generalizable neural
  radiance fields,'' in \emph{Proceedings of the IEEE/CVF Conference on
  Computer Vision and Pattern Recognition}, 2023, pp. 24--34.

\bibitem{teed2021droid}
Z.~Teed and J.~Deng, ``Droid-slam: Deep visual slam for monocular, stereo, and
  rgb-d cameras,'' \emph{Advances in neural information processing systems},
  vol.~34, pp. 16\,558--16\,569, 2021.

\bibitem{meuleman2023progressively}
A.~Meuleman, Y.-L. Liu, C.~Gao, J.-B. Huang, C.~Kim, M.~H. Kim, and J.~Kopf,
  ``Progressively optimized local radiance fields for robust view synthesis,''
  in \emph{Proceedings of the IEEE/CVF Conference on Computer Vision and
  Pattern Recognition}, 2023, pp. 16\,539--16\,548.

\bibitem{ran2024ct}
Y.~Ran, Y.~Li, Q.~Ye, Y.~Huo, Z.~Bai, J.~Sun, and J.~Chen, ``Ct-nerf:
  Incremental optimizing neural radiance field and poses with complex
  trajectory,'' \emph{arXiv preprint arXiv:2404.13896}, 2024.

\bibitem{edstedt2023dkm}
J.~Edstedt, I.~Athanasiadis, M.~Wadenb{\"a}ck, and M.~Felsberg, ``Dkm: Dense
  kernelized feature matching for geometry estimation,'' in \emph{Proceedings
  of the IEEE/CVF Conference on Computer Vision and Pattern Recognition}, 2023,
  pp. 17\,765--17\,775.

\bibitem{knapitsch2017tanks}
A.~Knapitsch, J.~Park, Q.-Y. Zhou, and V.~Koltun, ``Tanks and temples:
  Benchmarking large-scale scene reconstruction,'' \emph{ACM Transactions on
  Graphics (ToG)}, vol.~36, no.~4, pp. 1--13, 2017.

\bibitem{yu2021plenoctrees}
A.~Yu, R.~Li, M.~Tancik, H.~Li, R.~Ng, and A.~Kanazawa, ``Plenoctrees for
  real-time rendering of neural radiance fields,'' in \emph{Proceedings of the
  IEEE/CVF International Conference on Computer Vision}, 2021, pp. 5752--5761.

\bibitem{muller2022instant}
T.~M{\"u}ller, A.~Evans, C.~Schied, and A.~Keller, ``Instant neural graphics
  primitives with a multiresolution hash encoding,'' \emph{ACM Transactions on
  Graphics (ToG)}, vol.~41, no.~4, pp. 1--15, 2022.

\bibitem{barron2022mip}
J.~T. Barron, B.~Mildenhall, D.~Verbin, P.~P. Srinivasan, and P.~Hedman,
  ``Mip-nerf 360: Unbounded anti-aliased neural radiance fields,'' in
  \emph{Proceedings of the IEEE/CVF Conference on Computer Vision and Pattern
  Recognition}, 2022, pp. 5470--5479.

\bibitem{zhang2024fregs}
J.~Zhang, F.~Zhan, M.~Xu, S.~Lu, and E.~Xing, ``Fregs: 3d gaussian splatting
  with progressive frequency regularization,'' in \emph{Proceedings of the
  IEEE/CVF Conference on Computer Vision and Pattern Recognition}, 2024, pp.
  21\,424--21\,433.

\bibitem{fan2023lightgaussian}
Z.~Fan, K.~Wang, K.~Wen, Z.~Zhu, D.~Xu, and Z.~Wang, ``Lightgaussian: Unbounded
  3d gaussian compression with 15x reduction and 200+ fps,'' \emph{arXiv
  preprint arXiv:2311.17245}, 2023.

\bibitem{barron2021mip}
J.~T. Barron, B.~Mildenhall, M.~Tancik, P.~Hedman, R.~Martin-Brualla, and P.~P.
  Srinivasan, ``Mip-nerf: A multiscale representation for anti-aliasing neural
  radiance fields,'' in \emph{Proceedings of the IEEE/CVF International
  Conference on Computer Vision}, 2021, pp. 5855--5864.

\bibitem{yu2024mip}
Z.~Yu, A.~Chen, B.~Huang, T.~Sattler, and A.~Geiger, ``Mip-splatting:
  Alias-free 3d gaussian splatting,'' in \emph{Proceedings of the IEEE/CVF
  Conference on Computer Vision and Pattern Recognition}, 2024, pp.
  19\,447--19\,456.

\bibitem{jain2021putting}
A.~Jain, M.~Tancik, and P.~Abbeel, ``Putting nerf on a diet: Semantically
  consistent few-shot view synthesis,'' in \emph{Proceedings of the IEEE/CVF
  International Conference on Computer Vision}, 2021, pp. 5885--5894.

\bibitem{yang2023freenerf}
J.~Yang, M.~Pavone, and Y.~Wang, ``Freenerf: Improving few-shot neural
  rendering with free frequency regularization,'' in \emph{Proceedings of the
  IEEE/CVF conference on computer vision and pattern recognition}, 2023, pp.
  8254--8263.

\bibitem{yariv2021volume}
L.~Yariv, J.~Gu, Y.~Kasten, and Y.~Lipman, ``Volume rendering of neural
  implicit surfaces,'' \emph{Advances in Neural Information Processing
  Systems}, vol.~34, pp. 4805--4815, 2021.

\bibitem{wang2021neus}
P.~Wang, L.~Liu, Y.~Liu, C.~Theobalt, T.~Komura, and W.~Wang, ``Neus: Learning
  neural implicit surfaces by volume rendering for multi-view reconstruction,''
  \emph{arXiv preprint arXiv:2106.10689}, 2021.

\bibitem{guedon2024sugar}
A.~Gu{\'e}don and V.~Lepetit, ``Sugar: Surface-aligned gaussian splatting for
  efficient 3d mesh reconstruction and high-quality mesh rendering,'' in
  \emph{Proceedings of the IEEE/CVF Conference on Computer Vision and Pattern
  Recognition}, 2024, pp. 5354--5363.

\bibitem{fu2022geo}
Q.~Fu, Q.~Xu, Y.~S. Ong, and W.~Tao, ``Geo-neus: Geometry-consistent neural
  implicit surfaces learning for multi-view reconstruction,'' \emph{Advances in
  Neural Information Processing Systems}, vol.~35, pp. 3403--3416, 2022.

\bibitem{yen2021inerf}
L.~Yen-Chen, P.~Florence, J.~T. Barron, A.~Rodriguez, P.~Isola, and T.-Y. Lin,
  ``inerf: Inverting neural radiance fields for pose estimation,'' in
  \emph{2021 IEEE/RSJ International Conference on Intelligent Robots and
  Systems (IROS)}.\hskip 1em plus 0.5em minus 0.4em\relax IEEE, 2021, pp.
  1323--1330.

\bibitem{boss2022samurai}
M.~Boss, A.~Engelhardt, A.~Kar, Y.~Li, D.~Sun, J.~T. Barron, H.~Lensch, and
  V.~Jampani, ``Samurai: Shape and material from unconstrained real-world
  arbitrary image collections,'' \emph{arXiv preprint arXiv:2205.15768}, 2022.

\bibitem{tian2023mononerf}
F.~Tian, S.~Du, and Y.~Duan, ``Mononerf: Learning a generalizable dynamic
  radiance field from monocular videos,'' in \emph{Proceedings of the IEEE/CVF
  International Conference on Computer Vision}, 2023, pp. 17\,903--17\,913.

\bibitem{hartley1997defense}
R.~I. Hartley, ``In defense of the eight-point algorithm,'' \emph{IEEE
  Transactions on pattern analysis and machine intelligence}, vol.~19, no.~6,
  pp. 580--593, 1997.

\bibitem{fathy2011fundamental}
M.~E. Fathy, A.~S. Hussein, and M.~F. Tolba, ``Fundamental matrix estimation: A
  study of error criteria,'' \emph{Pattern Recognition Letters}, vol.~32,
  no.~2, pp. 383--391, 2011.

\bibitem{hartley2003multiple}
R.~Hartley and A.~Zisserman, \emph{Multiple view geometry in computer
  vision}.\hskip 1em plus 0.5em minus 0.4em\relax Cambridge university press,
  2003.

\bibitem{chen2022projective}
J.~Chen, Y.~Yin, T.~Birdal, B.~Chen, L.~J. Guibas, and H.~Wang, ``Projective
  manifold gradient layer for deep rotation regression,'' in \emph{Proceedings
  of the IEEE/CVF Conference on Computer Vision and Pattern Recognition}, 2022,
  pp. 6646--6655.

\bibitem{kingma2014adam}
D.~P. Kingma and J.~Ba, ``Adam: A method for stochastic optimization,''
  \emph{arXiv preprint arXiv:1412.6980}, 2014.

\bibitem{li2022nerfacc}
R.~Li, M.~Tancik, and A.~Kanazawa, ``Nerfacc: A general nerf acceleration
  toolbox,'' \emph{arXiv preprint arXiv:2210.04847}, 2022.

\bibitem{wang2024dust3r}
S.~Wang, V.~Leroy, Y.~Cabon, B.~Chidlovskii, and J.~Revaud, ``Dust3r: Geometric
  3d vision made easy,'' in \emph{Proceedings of the IEEE/CVF Conference on
  Computer Vision and Pattern Recognition}, 2024, pp. 20\,697--20\,709.

\end{thebibliography}

\end{document}